\documentclass[10pt,twocolumn,letterpaper]{article}

\usepackage{cvpr} 
\usepackage{multirow}
\usepackage{booktabs}
\usepackage{pifont}

\usepackage{graphicx}
\usepackage{amsmath}
\usepackage{amssymb}
\usepackage{booktabs}
\usepackage{blindtext}

\def\paperID{4545} 
\def\confName{CVPR}
\def\confYear{2025}

\title{Depth-Guided Bundle Sampling for Efficient Generalizable Neural Radiance Field Reconstruction\thanks{This work is supported by National Key R\&D Program of China under Grant No. 2021YFF0900701, and the National Natural Science Foundation of China under Grant No. 62001432.}}

\author{
Li Fang$^{1}$ \quad
Hao Zhu$^{2}$ \quad
Longlong Chen$^{1}$ \quad
Fei Hu$^{1}$\thanks{Corresponding author.} \quad
Long Ye$^{1}$ \quad
Zhan Ma$^{2}$ \\
$^{1}$Key Laboratory of Media Audio and Video (Communication University of China), \\
Ministry of Education, Beijing 100024, China \\
$^{2}$School of Electronic Science and Engineering, Nanjing University, Nanjing 210023, China \\
{\tt\small \{lifang8902, chenll, hufei, yelong\}@cuc.edu.cn} \quad
{\tt\small \{zhuhao\_photo, mazhan\}@nju.edu.cn}
}

\begin{document}
\maketitle
\begin{abstract}

Recent advancements in generalizable novel view synthesis have achieved impressive quality through interpolation between nearby views. However, rendering high-resolution images remains computationally intensive due to the need for dense sampling of all rays. Recognizing that natural scenes are typically piecewise smooth and sampling all rays is often redundant, we propose a novel depth-guided bundle sampling strategy to accelerate rendering. By grouping adjacent rays into a bundle and sampling them collectively, a shared representation is generated for decoding all rays within the bundle. To further optimize efficiency, our adaptive sampling strategy dynamically allocates samples based on depth confidence, concentrating more samples in complex regions while reducing them in smoother areas. When applied to ENeRF, our method achieves up to a 1.27 dB PSNR improvement and a 47\% increase in FPS on the DTU dataset. Extensive experiments on synthetic and real-world datasets demonstrate state-of-the-art rendering quality and up to $2 \times$ faster rendering compared to existing generalizable methods. Code is available at \href{https://github.com/KLMAV-CUC/GDB-NeRF}{https://github.com/KLMAV-CUC/GDB-NeRF}.

\end{abstract}
    
\section{Introduction}
\label{sec:intro}

\begin{figure}[t]
  \centering
   \includegraphics[width=0.95\linewidth]{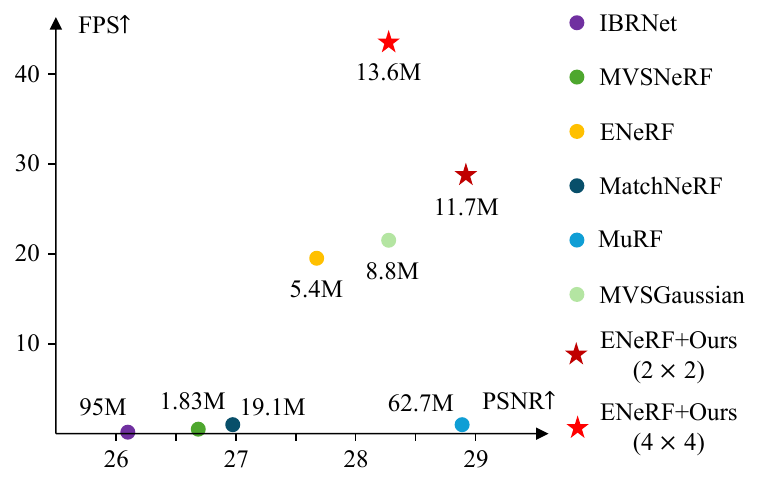}

   \caption{Rendering Quality (PSNR) vs. Speed (FPS) on the DTU Dataset \cite{aanaes2016large} under 3-view setting. ENeRF+Ours achieves state-of-the-art rendering quality and faster rendering speed, outperforming existing generalizable novel view synthesis methods \cite{wang2021ibrnet, chen2021mvsnerf, lin2022efficient, chen2023matchnerf, xu2024murf, liu2024mvsgaussian}. Model parameter counts are also provided.}
   \label{fig:teaser}
\end{figure}

Novel view synthesis aims to generate photorealistic images from new perspectives using a sparse set of input images. The advent of Neural Radiance Field (NeRF) \cite{mildenhall2020nerf} and subsequent advancements \cite{sun2024recent} have significantly improved the quality of view synthesis, garnering substantial interest in both academia and industry. NeRF models scenes as continuous radiance and density fields, producing high-quality images. However, this comes at a cost: NeRF's volume rendering process requires numerous samples per ray, leading to significant computational demands and impeding real-time, high-resolution synthesis. Additionally, NeRF models are typically scene-specific, requiring lengthy training for each new scene.

Recent efforts to improve NeRF efficiency have focused on compact data structures and lightweight MLP decoders to store radiance fields, significantly reducing both training and rendering times for scene-specific NeRFs \cite{muller2022instant, fridovich2022plenoxels, chen2023mobilenerf, hu2022efficientnerf, wang2023adaptive}. Techniques like baking and empty space skipping have also contributed to these improvements. However, these methods require per-scene optimization, limiting their ability to generalize to novel scenes. Generalizable NeRFs address this limitation by predicting radiance fields from multi-view images, enabling novel view synthesis for previously unseen scenes \cite{yu2021pixelnerf, wang2021ibrnet, chibane2021stereo, trevithick2021grf, li2021mine, chen2021mvsnerf, johari2022geonerf, lin2022efficient, liu2022neural, nguyen2023cascaded, wang2023attention, xu2024murf, tanay2024global, liu2024geometry}. Although promising, generalizable methods often face slower rendering speeds, as scene-specific optimizations are incompatible. While some approaches attempt to reduce per-ray sampling \cite{lin2022efficient, neff2021donerf}, real-time rendering remains challenging.

The expensive computational cost in NeRF-related methods is primarily due to the need for dense sampling across all pixels. From a plenoptic sampling perspective \cite{chai2000plenoptic}, this full-pixel sampling implicitly assumes high-frequency information is uniform across the scene. However, natural scenes are usually piecewise smooth, with high-frequency details in specific areas like object boundaries and depth discontinuities. Based on this observation, we propose a novel bundle sampling strategy. This method groups adjacent rays and samples them jointly at a global, low sampling rate for smooth regions. In non-smooth areas, high-frequency information is captured by increasing samples along the depth dimension. Furthermore, samples within each bundle are further constrained to a small range using depth maps. As a result, this approach significantly reduces the overall number of samples, greatly improving efficiency.


We integrate this depth-guided bundle sampling strategy into a representative generalizable NeRF, ENeRF \cite{lin2022efficient}. Instead of sampling points along individual rays, our method samples 3D areas along bundles of adjacent rays by casting a cone that encompasses the corresponding pixels. Within this cone, inscribed spheres are sampled and featurized with pre-filtered features extracted from the source views. Through depth-guided adaptive sampling, sampling is constrained to a predicted depth range, with the number of samples adjusted dynamically: in smooth regions with a narrow depth range, fewer samples suffice for accurate reconstruction, while areas with a broader depth range require more samples to capture finer details. This adaptive approach aligns with the plenoptic principle \cite{chai2000plenoptic}, ensuring each sample covers a narrow depth range near surfaces. A bundle representation is then computed via volume rendering and decoded to predict the color of each ray. We also apply this depth-guided bundle sampling strategy to a recent generalizable 3D Gaussian splatting (3D-GS) method \cite{liu2024mvsgaussian}. Extensive experiments demonstrate that our method significantly improves the efficiency of base models and achieves synthesis quality comparable to state-of-the-art generalizable models, with significant improvements in rendering speed.

Our main contributions can be summarized as follows:
\begin{itemize}
  \item We propose a novel bundle sampling strategy that jointly samples bundles of adjacent rays, significantly reducing the number of samples needed and improving both rendering speed and quality.
  \item We introduce a depth-guided adaptive sampling method that dynamically adjusts the number of samples based on local depth range, achieving over a 50\% increase in rendering speed without sacrificing quality.
  \item We validate our method on both generalizable NeRF and 3D-GS across various datasets. Our method achieves state-of-the-art rendering quality and up to $2 \times$ faster rendering than existing generalizable NeRFs. It also offers flexibility, allowing adjustments between quality and efficiency depending on practical requirements.
\end{itemize}

\section{Related Works}
\label{sec:rw}
\paragraph{Accelerating NeRF.} NeRFs represent scenes as continuous color and density fields using neural networks, achieving high-quality view synthesis via volume rendering. Despite their success, NeRFs suffer from high computational demands, slowing training and inference. To address this, recent methods have replaced large neural representations with local sparse feature fields and shallow MLP decoders \cite{liu2020neural, yu2021plenoctrees, sun2022direct, muller2022instant, chen2022tensorf}, or spherical harmonics embeddings \cite{fridovich2022plenoxels, chen2022tensorf, karnewar2022relu}. Other solutions utilize voxel grids \cite{liu2020neural}, octrees \cite{fridovich2022plenoxels}, tensor decompositions \cite{chen2022tensorf}, or 3D Gaussians \cite{kerbl20233d}, as well as efficient sampling \cite{neff2021donerf, hu2022efficientnerf, kurz2022adanerf, lin2022efficient} and image-space convolutions \cite{cao2023real, wan2023learning}, leading to significant speed-ups. However, these improvements remain scene-specific, requiring training for each new scene.

\paragraph{Generalizable NeRF.} These methods generalize NeRF by encoding features for each 3D point and decoding them into volume density and radiance, akin to image-based rendering through NeRF-like volume rendering techniques \cite{yu2021pixelnerf, wang2021ibrnet, chibane2021stereo, trevithick2021grf, li2021mine, chen2021mvsnerf, johari2022geonerf, lin2022efficient, liu2022neural, chen2023matchnerf, nguyen2023cascaded, wang2023attention, tanay2024global, liu2024geometry}. While these methods improve across-scene generalizability, they remain limited by slow optimization and rendering. CG-NeRF \cite{nguyen2023cascaded} predicts coarse radiance fields and depth, refining fine-scale features sampled near the estimated depth. ConvGLR \cite{tanay2024global} processes all camera rays in a low-resolution latent space, which improves quality in sparse-view synthesis. GeFu \cite{liu2024geometry} improves generalizable NeRFs through geometry-aware reconstruction with Adaptive Cost Aggregation and Consistency-Aware Fusion. Most of these methods lack real-time rendering capability. ENeRF \cite{lin2022efficient} accelerates generalizable NeRFs with depth-guided sampling but uses a fixed number of samples per ray, leading to inefficiency. In contrast, our method adaptively samples ray bundles based on local scene complexity, providing greater flexibility and efficiency.

\paragraph{Generalizable 3D-GS.} Recent advancements in generalizable Gaussian splatting methods have significantly enhanced the efficiency of novel view synthesis. These approaches eliminate the need for per-scene optimization by directly regressing Gaussian parameters through a feed-forward architecture. PixelSplat \cite{charatan2024pixelsplat} tackles scale ambiguity by utilizing a multiview epipolar transformer and scale-aware features to predict pixel-aligned 3D Gaussian primitives, though its design is limited to image pair inputs. GPS-Gaussian \cite{zheng2024gps} extends this approach to stereo matching, incorporating epipolar rectification, disparity estimation, and feature encoding. However, its reliance on ground-truth depth maps restricts its applicability. Splatter Image \cite{szymanowicz2024splatter} proposes a single-view reconstruction approach based on Gaussian splatting, though its focus on object-centric reconstruction constrains its generalization to unseen scenes. MVSGaussian \cite{liu2024mvsgaussian} combines multi-view stereo with a generalizable 3D Gaussian representation for real-time novel view synthesis. However, its single-sample-per-ray strategy necessitates highly accurate depth estimations and hinders further efficiency improvements.

\section{Background}
\paragraph{Preliminaries on Generalizable NeRF.} Given input multi-view images with intrinsic and extrinsic camera matrices, NeRF constructs the scene’s radiance fields using a coordinate-based network. However, optimizing this network for each individual scene is computationally intensive. To address this, generalizable NeRF incorporates an additional geometry-aware feature, $\boldsymbol{f}_p$, extracted from the source images and included in the network input:
\begin{equation}
    \sigma_p, c_p = NN_{\theta}(\boldsymbol{x}_p, \boldsymbol{d}_p, \boldsymbol{f}_p),
\label{eq:gnerf}
\end{equation}
where $\boldsymbol{x}_p$ and $\boldsymbol{d}_p$ represent the coordinates and viewing direction of the 3D point $p$, respectively. Here, $\sigma_p$ and $c_p$ denote the density and color of $p$, defining the radiance field properties at this point. The coordinate network, $NN_{\theta}$, decodes these radiance field attributes. Once the radiance fields for the entire scene are constructed, volume rendering techniques can synthesize the color for a pixel from any desired viewpoint.

For an image of resolution $H \times W$, NeRF samples $N$ points along the ray passing through each pixel, leading to a computational complexity of approximately $\mathcal{O}(HWN)$. As the number of samples increases, so does the computational burden, posing challenges for rendering efficiency and motivating the need for optimized sampling and rendering strategies.

\paragraph{Plenoptic Sampling Theory.} To enhance efficiency, existing generalizable NeRFs aim to reduce the number of samples along each ray by concentrating sampling near scene surfaces \cite{lin2022efficient, liu2024mvsgaussian, liu2024geometry}. However, these methods do not explore the reduction of the total number of rays required for rendering; instead, they uniformly sample all rays with a fixed number of samples. Some approaches attempt to reduce ray sampling needs through super-resolution techniques, though these are often constrained to specific content types such as faces \cite{chan2022efficient}.

Plenoptic sampling theory, introduced in \cite{chai2000plenoptic}, investigates the sampling bounds in light field rendering to determine the minimum angular density required for accurate reconstruction. The theory defines the maximum allowable camera spacing as:
\begin{equation}
    \Delta t \le \frac{\text{max} \left( 2 \Delta v, 1 / B_v^s \right)}{f \cdot h_d}, 
\label{eq:max_spacing}
\end{equation}
where $f$ is the focal length, $\Delta v$ is the spacing between camera rays, $B_v^s$ is the highest texture frequency, and $h_d$ is the disparity range defined as $h_d = \frac{1}{z_\text{min}} - \frac{1}{z_\text{max}}$, with $z_\text{min}$ and $z_\text{max}$ denoting the minimum and maximum scene depths, respectively. \cref{eq:max_spacing} first implies that the ray spacing $\Delta v$ must satisfy the Nyquist sampling criterion, \textit{i.e.}, $\Delta v \le \frac{1}{2B_v^s}$, to avoid aliasing; it then highlights that geometric accuracy affects the sampling rate required for image capture: by decomposing the scene into $D$ depth ranges and sampling the light field in each range separately, the camera sampling interval can be increased by a factor of $D$.

Considering the relation between light fields and NeRFs \cite{ramamoorthi2024sampling}, plenoptic sampling theory inspires our depth-guided bundle sampling strategy. On the one hand, the piecewise smooth nature of natural scenes means that high-frequency content is localized, with most regions being smooth. This allows for reduced sampling rates in smooth areas by effectively lowering the camera resolution (\textit{i.e.}, increasing $\Delta v$). However, this may cause aliasing in high-frequency regions (\textit{e.g.}, edges, textures, thin structures). On the other hand, given a fixed camera spacing $\Delta t$, each sample in NeRF can be regarded as a local light field confined to a small disparity range, $h_d \le \frac{\text{max}\left( 2 \Delta v, \frac{1}{B_v^s} \right)}{f \cdot \Delta t}$. For high-frequency regions, they often suffer from inaccurate depth estimation, necessitating more samples to ensure each sample covers a small disparity range. These observations motivate our depth-guided bundle sampling strategy, which adapts the sampling density based on scene complexity.

\section{Method}
\begin{figure*}[t]
  \centering
  \includegraphics[width=0.95\linewidth]{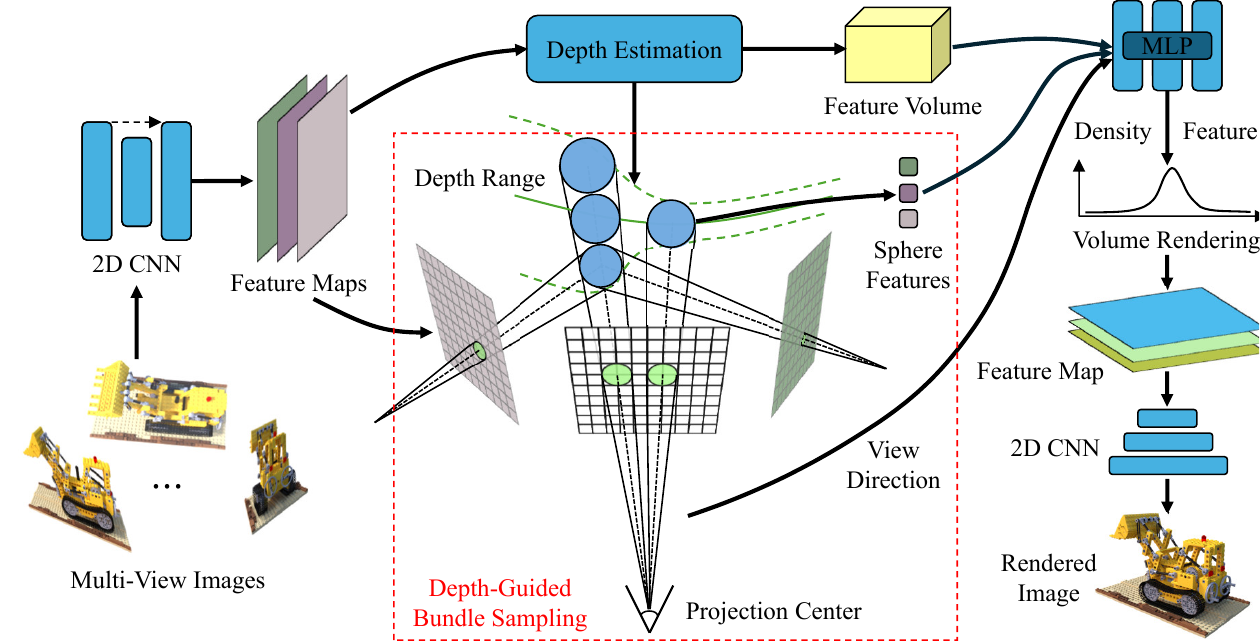}
  \caption{Network architecture of the proposed depth-guided bundle sampling strategy on ENeRF \cite{lin2022efficient}, denoted as ENeRF+Ours, which consists of four main components: (1) Multi-scale feature extraction; (2) Depth estimation to predict depth range; (3) Depth-guided bundle sampling, where rays are grouped into bundles and sampled adaptively based on predicted depth confidence; and (4) Radiance field prediction, which decodes each bundle's representation into individual ray colors.}
  \label{fig:pipeline}
\end{figure*}

Our method is designed to synthesize target views from $V$ nearby source views $\{ \boldsymbol{I}_i \}_{i=1}^V$ at interactive frame rates, addressing challenges in efficient sampling. The proposed depth-guided bundle sampling strategy includes two main innovations: a novel bundle sampling approach that groups adjacent rays in the target view for joint sampling, and a depth-guided adaptive sampling method that dynamically allocates samples based on depth confidence. This adaptive approach reduces the overall number of samples while maintaining consistency with the plenoptic sampling principles. We apply our method to representative ENeRF \cite{lin2022efficient} and MVSGaussian \cite{liu2024mvsgaussian} to demonstrate its effectiveness in generalizable NeRF and 3D-GS, respectively. \cref{fig:pipeline} presents the architecture of ENeRF+Ours, and a similar architecture is employed in MVSGaussian, where the sample features are also decoded into 3D-GS parameters.

\subsection{Bundle Sampling}
\label{sec:gdb}
\begin{figure}[t]
  \centering
  \subfloat[Ray Sampling]{
    \includegraphics[width=0.95\columnwidth]{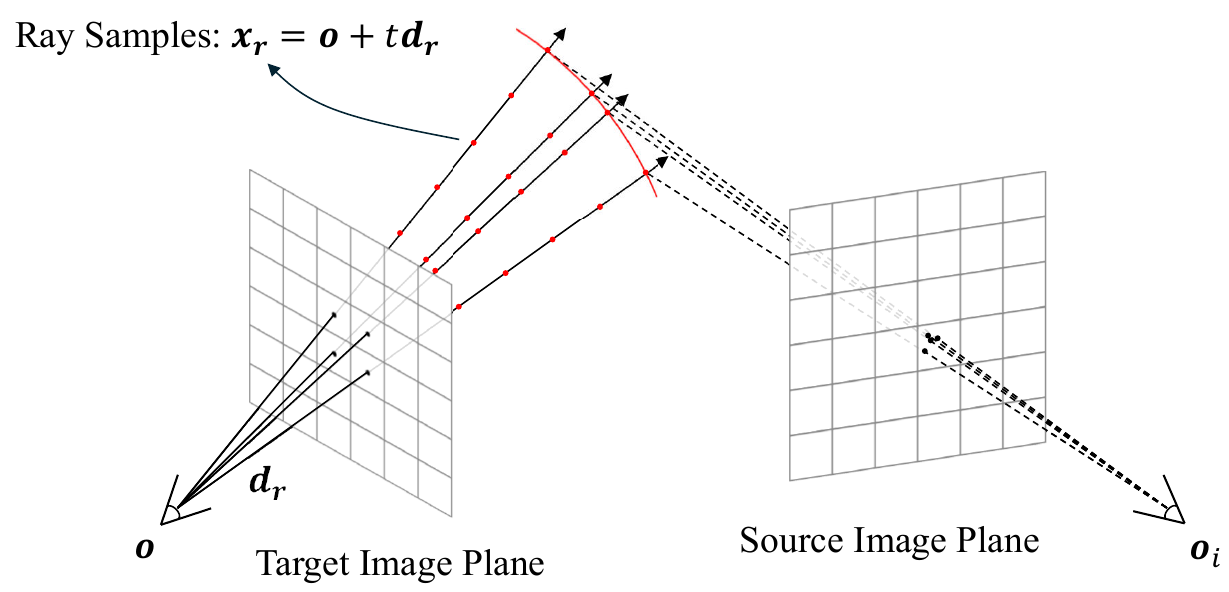}
  }
  \quad
  \subfloat[Bundle Sampling]{
    \includegraphics[width=0.95\columnwidth]{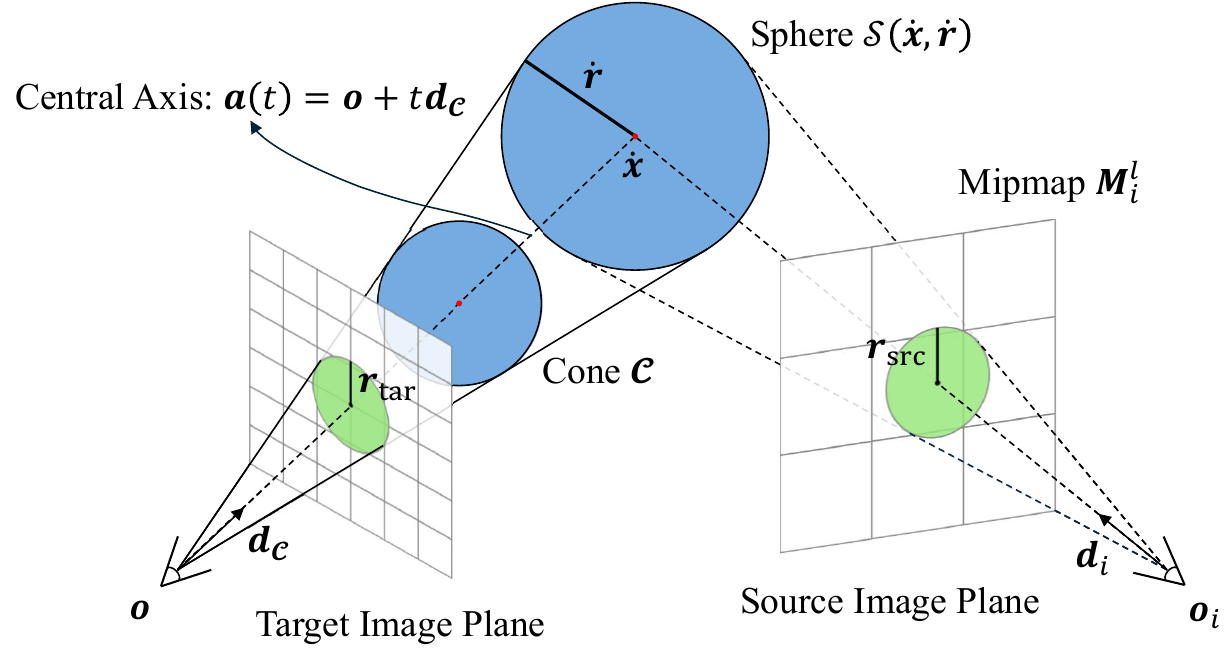}
  }
  \caption{Comparison of ray sampling strategies: (a) Traditional ray sampling, which processes each ray individually; (b) Proposed bundle sampling, which jointly samples neighboring rays to leverage scene coherence.}
  \label{fig:bundle_sampling}
\end{figure}

Our bundle sampling method improves efficiency by reducing the number of sampled rays. Adjacent rays in the target view are grouped into bundles, modeled as cones, with inscribed spheres sampled within each cone, as illustrated in \cref{fig:bundle_sampling}. This design broadens the angular coverage per sample, effectively reducing the total number of rays required. A similar sphere-based sampling approach is used in \cite{hu2023tri}, where spheres are cast for each ray’s cone to achieve anti-aliasing. In contrast, our method casts spheres for a cone of grouped rays, enabling collective sampling across multiple rays rather than treating them individually. However, this expanded coverage inevitably leads to loss of high-frequency details in the rendered image. To mitigate this, we encode each sample with both a joint bundle representation and a ray-specific representation, ensuring efficiency while preserving the necessary level of detail.

\subsubsection{Sphere-based cone sampling}
Our method models bundles of rays as cones passing through groups of adjacent pixels in the target view. We begin by casting rays from the camera's projection center $\boldsymbol{o}$ through each pixel center $\boldsymbol{p}_{\boldsymbol{o}}$, with ray directions defined as $\boldsymbol{d}_{\boldsymbol{r}} = \boldsymbol{p}_{\boldsymbol{o}} - \boldsymbol{o}$. Points along each ray are parameterized as $\boldsymbol{x}_{\boldsymbol{r}} = \boldsymbol{o} + t \boldsymbol{d}_{\boldsymbol{r}}$, where $t$ is the distance along the ray.

As shown in \cref{fig:bundle_sampling}(b), for a bundle of $K \times K$ rays, the corresponding region on the target image plane is modeled as a disk with radius $r_\text{tar} = K \cdot r_p$, where $r_p = \sqrt{\Delta x \cdot \Delta y / \pi}$ is the radius of a single-pixel disk, and $\Delta x$ and $\Delta y$ are the pixel's width and height in world coordinates, derived from camera parameters. The cone $\mathcal{C}$ is emitted from the camera's projection center along the average direction $\boldsymbol{d}_{\mathcal{C}} = \sum_{\boldsymbol{r} \in \mathcal{C}} \boldsymbol{d}_{\boldsymbol{r}} / K^2$, and intersects the image plane at the disk, with the cone's axis defined as $\boldsymbol{a}(t) = \boldsymbol{o} + t \boldsymbol{d}_{\mathcal{C}}$. The cone is sampled using a set of inscribed spheres, $\mathcal{S} (\dot{\boldsymbol{x}}, \dot{r})$, defined by their centers $\dot{\boldsymbol{x}}$ and radii $\dot{r}$ as follows:
\begin{equation}
\begin{split}
    & \dot{\boldsymbol{x}} = \frac{1}{K^2} \sum_{\boldsymbol{r} \in \mathcal{C}} \boldsymbol{x}_{\boldsymbol{r}}, \\
    & \dot{r} = \frac{{\left\| \dot{\boldsymbol{x}} - \boldsymbol{o} \right\|}_2 \cdot f \cdot r_\text{tar}}{{\left\| \boldsymbol{d}_{\mathcal{C}} \right\|}_2 \cdot \sqrt{{\left( \sqrt{{\left\| \boldsymbol{d}_{\mathcal{C}} \right\|}_2^2 - f^2} - r_\text{tar} \right)}^2 + f^2}}.
\end{split}
\label{eq:sphere-radius}
\end{equation}
Given sorted distances $\{ t_n \}_{n=1}^V$, the sphere centers $\dot{\boldsymbol{x}}_n$ and radii $\dot{r}_n$ are computed.

For a target view with resolution $H \times W$, the image is divided into $\frac{H}{K} \times \frac{W}{K}$ bundles, each consisting of $K \times K$ pixels. This reduces the total number of samples by approximately $1 / K^2$, ensuring efficient angular coverage without sacrificing essential surface detail.

\subsubsection{Multi-view image-based sphere encoding}
Each sphere covers a broad angular range, which extends to its feature representation. As shown in \cref{fig:bundle_sampling}(b), we extract features based on the area occupied by each sphere, forming a joint bundle representation. However, this approach may omit high-frequency details. To retain fine-grained information from individual rays, we augment the joint bundle representation with a ray-specific representation.

The joint bundle representation is derived from the area occupied by the spheres, necessitating an adjustment of the source view pixel spacing $\Delta v$ in \cref{eq:max_spacing}. Inspired by area-sampling techniques used in scene-specific anti-aliasing NeRFs \cite{barron2021mip, barron2022mip, hu2023tri}, we incorporate mipmap representations from \cite{hu2023tri} into a multi-view image-based encoding framework. We extract features $\{ \boldsymbol{F}_i \}_{i=1}^V$ from source views and convert them into mipmaps $\{ \boldsymbol{M}_i \}_{i=1}^V$, creating a hierarchical structure that represents the pre-filtered feature space. The base level $\boldsymbol{M}_i^{l_0}$ is the original feature map $\boldsymbol{F}_i$, with each subsequent level reducing the spatial resolution by a factor of two. To compute the feature vector $\{ \boldsymbol{f}_{\mathcal{S}, i} \}_{i=1}^V$ for sphere $\mathcal{S} (\dot{\boldsymbol{x}}, \dot{r})$, we project it onto the $i$-th source view to create a disc, with its radius $r_\text{src}$ calculated as:
\begin{equation}
    r_\text{src} = \frac{{\left\| \boldsymbol{d}_i \right\|}_2^2}{f \cdot \sqrt{{\left( \frac{{\left\| \dot{\boldsymbol{x}} - \boldsymbol{o}_i \right\|}_2}{\dot{r}} \right)}^2 - 1} + \sqrt{{\left\| \boldsymbol{d}_i \right\|}_2^2 - f^2}}, 
\label{eq:source-radius}
\end{equation}
where $\boldsymbol{o}_i$ is the camera's projection center of the $i$-th source view, and ${\left\| \boldsymbol{d}_i \right\|}_2$ is the distance from the projection of $\dot{\boldsymbol{x}}$ onto the image plane to $\boldsymbol{o}_i$ (see \cref{fig:bundle_sampling}(b)). The appropriate mipmap level $l$ corresponding to the sphere is determined using:
\begin{equation}
    l = \log_2 \left( \frac{r_\text{src}}{r_p} \right). 
\label{eq:mip-level}
\end{equation}
This process aligns the radius $r_\text{src}$ of the sphere's projection with the corresponding feature elements’ radius at a specific level $l$ within $\boldsymbol{M}_i^l$, which is equivalent to adjusting the spacing $\Delta v$ of the sampling camera. Once the appropriate mipmap level is identified, the feature vector $\boldsymbol{f}_{\mathcal{S}, i}$ can be accurately extracted from the mipmap $\boldsymbol{M}_i$ via trilinear interpolation at $(x, y, l)$. 

For ray-specific representation, the 3D points $\boldsymbol{x}_{\boldsymbol{r}}$ associated with the sphere $\mathcal{S} (\dot{\boldsymbol{x}}, \dot{r})$ are projected onto the source views $\{ \boldsymbol{I}_i \}_{i=1}^V$ to extract pixel-aligned colors. The extracted colors corresponding to the $K \times K$ rays within each bundle are concatenated to form the ray-specific representation, denoted as $\{ \boldsymbol{f}_{\boldsymbol{r}, i} \}_{i=1}^V$, which captures high-frequency information for the rays as projected onto source views.

By combining the joint bundle representation and ray-specific representation, our approach efficiently encodes the bundle samples while maintaining necessary scene details for accurate reconstruction.

\subsection{Depth-guided Adaptive Sampling}
\label{sec:depth-guided-ada}
To further optimize sampling, we restrict it to a predicted depth range, $\boldsymbol{R}$, estimated based on the depth value and associated confidence interval \cite{lin2022efficient, liu2024geometry, liu2024mvsgaussian}. According to \cref{eq:max_spacing}, for a fixed camera spacing $\Delta t$, the disparity range covered by the local light field of each sample must satisfy $h_d \le \frac{\text{max} \left( 2 \Delta v, 1 / B_v^s \right)}{f \cdot \Delta t}$, representing the maximum sampling spacing along the depth. Consequently, in smooth regions with a narrow depth range, fewer samples are sufficient for accurate reconstruction, while regions with a larger depth range require additional samples. Our depth-guided adaptive sampling strategy dynamically allocates samples near surfaces, ensuring that each sample covers a narrow depth range. The number of samples per bundle is computed as:
\begin{equation}
    N_\mathcal{C} = \text{max} \left( \text{ceil} \left( 2 \cdot \boldsymbol{R} / \delta_s \right), N_{\text{max}} \right), 
\label{eq:num-sample}
\end{equation}
where $\delta_s$ denotes the minimum inter-sample spacing, with a maximum limit of $N_{\text{max}}$ samples. The depth range is then divided into $N_\mathcal{C}$ uniform intervals, with sampling performed at the center of each interval. For cases requiring only one sample, the sample is placed at the predicted surface point.

Unlike existing depth-guided methods that apply uniform ray sampling with a fixed number of points, our adaptive approach dynamically adjusts the number of samples along the depth according to plenoptic sampling theory, significantly enhancing both efficiency and adaptability.

\subsection{Radiance Field Prediction}
\label{sec:nerf}
For each sampled sphere, we predict the density $\sigma$ and blending weights $\{ w_i \}_{i=1}^V$ as follows:
\begin{equation}
    \sigma, \{ w_i \}_{i=1}^V = \text{MLP} \left( \{ \boldsymbol{f}_{\mathcal{S}, i} \}_{i=1}^V, \boldsymbol{f}_{\text{voxel}}, \{ \Delta \boldsymbol{d}_i \}_{i=1}^V \right), 
\label{eq:nerf-mlp}
\end{equation}
where $\boldsymbol{f}_{\text{voxel}}$ is the voxel-aligned feature extracted from the feature volume (from depth estimation, see \cref{fig:pipeline}) by applying trilinear interpolation at the sphere center $\dot{\boldsymbol{x}}$. $\Delta \boldsymbol{d}_i$ is the concatenation of the norm and direction of $\boldsymbol{d}_i - \boldsymbol{d}_{\mathcal{C}}$, with $\boldsymbol{d}_i$ as the direction of $\dot{\boldsymbol{x}}$ in the $i$-th source view. The network design is adapted from \cite{lin2022efficient}. The blending weights $\{ w_i \}_{i=1}^V$ combine the features using softmax:
\begin{equation}
    \{ \boldsymbol{f}_{\mathcal{S}}, \boldsymbol{f}_{\boldsymbol{r}} \} = \sum_{i=1}^V \frac{\text{exp} (w_i) \cdot \{ \boldsymbol{f}_{\mathcal{S}, i} \boldsymbol{f}_{\boldsymbol{r}, i} \}}{\sum_{j=1}^V \text{exp} (w_j)}.
\label{eq:blending}
\end{equation}

We employ volume rendering to compute the per-bundle features $\{ \boldsymbol{F}_{\mathcal{C}}, \boldsymbol{F}_{\boldsymbol{r}} \}$ by accumulating the features $\{ \boldsymbol{f}_{\mathcal{S}}, \boldsymbol{f}_{\boldsymbol{r}} \}_n$ across the $N_{\mathcal{C}}$ sampled spheres along the cone $\mathcal{C}$:
\begin{equation}
    \{ \boldsymbol{F}_{\mathcal{C}}, \boldsymbol{F}_{\boldsymbol{r}} \} = \sum_{n=1}^{N_{\mathcal{C}}} \tau_n \left( 1 - \text{exp} (-\sigma_n) \right) \{ \boldsymbol{f}_{\mathcal{S}}, \boldsymbol{f}_{\boldsymbol{r}} \}_n,
\label{eq:accumulate}
\end{equation}
where $\tau_n$ is the accumulated volume transmittance from the cone apex to the sphere center, computed as:
\begin{equation}
    \tau_n = \text{exp} (-\sum_{j=1}^{n-1} \sigma_j).
\label{eq:transmittance}
\end{equation}

\textit{Neural renderer.} For the target view with spatial dimensions $H \times W$, the image is divided into $\frac{H}{K} \times \frac{W}{K}$ non-overlapping bundles, each containing $K \times K$ rays. Using \cref{eq:accumulate}, we generate a feature map $\{ \boldsymbol{F}_{\mathcal{C}}^i, \boldsymbol{F}_{\boldsymbol{r}}^i \}_{i=1}^{\frac{H}{K} \times \frac{W}{K}}$. Here, $\boldsymbol{F}_{\mathcal{C}}^i$ represents the aggregated joint bundle representation, which is processed through several residual convolutional blocks and decoded into coarse ray colors via sub-pixel convolution, yielding a coarse RGB map $\boldsymbol{\hat{I}}_c$ of dimensions $H \times W \times 3$. The second component, $\boldsymbol{F}_{\boldsymbol{r}}^i$, contains the folded RGB values for individual rays and is unfolded to generate a fine RGB map $\boldsymbol{\hat{I}}_f$ of dimensions $H \times W \times 3$, preserving high-frequency details. The final high-resolution target view is obtained by summing these two components:
\begin{equation}
    \boldsymbol{\hat{I}}_{\text{tar}} = \boldsymbol{\hat{I}}_c + \boldsymbol{\hat{I}}_f.
\label{eq:renderer}
\end{equation}

For application to MVSGaussian \cite{liu2024mvsgaussian}, the renderer not only outputs the rendered target view image but also produces a feature map, which is subsequently processed to generate 3D-GS parameters following \cite{liu2024mvsgaussian}.
\section{Experiments}
\subsection{Settings}

\begin{table*}[t]
\small
\centering
\begin{tabular}{ccccccccc}
\hline
\multirow{2}{*}{Methods} & \multicolumn{5}{c}{3-view} & \multicolumn{3}{c}{2-view} \\
\cmidrule(r){2-6} \cmidrule(r){7-9}
 & PSNR $\uparrow$ & SSIM $\uparrow$ & LPIPS $\downarrow$ & Avg. samples per ray & FPS $\uparrow$ & PSNR $\uparrow$ & SSIM $\uparrow$ & LPIPS $\downarrow$ \\
\hline
PixelNeRF \cite{yu2021pixelnerf} & 19.31 & 0.789 & 0.382 & 96 & 0.019 & - & - & - \\
IBRNet \cite{wang2021ibrnet} & 26.04 & 0.917 & 0.191 & 128 & 0.217 & - & - & - \\
MVSNeRF \cite{chen2021mvsnerf} & 26.63 & 0.931 & 0.168 & 128 & 0.416 & 24.03 & 0.914 & 0.192 \\
ENeRF \cite{lin2022efficient} & 27.61 & 0.957 & 0.089 & 2 & 19.5 & 25.48 & 0.942 & 0.107 \\
MatchNeRF \cite{chen2023matchnerf} & 26.91 & 0.934 & 0.159 & 128 & 1.04 & 25.03 & 0.919 & 0.181 \\
GNT \cite{wang2023attention} & 26.39 & 0.923 & 0.156 & 192 & 0.01 & 24.32 & 0.903 & 0.201 \\
CG-NeRF \cite{nguyen2023cascaded} & 28.21 & 0.930 & 0.170 & 4 & 2.56 & - & - & - \\
MuRF \cite{xu2024murf} & 28.76 & 0.961 & 0.077 & 80 & 0.934 & 25.61 & 0.938 & 0.104 \\
ConvGLR \cite{tanay2024global} & \textbf{31.65} & 0.952 & 0.080 & 128 & 0.825 & - & - & - \\
MVSGaussian \cite{liu2024mvsgaussian} & 28.21 & \underline{0.963} & \underline{0.076} & 1 & 21.5 & 25.78 & \underline{0.947} & 0.095 \\
ENeRF+Ours ($2 \times 2$) & \underline{28.86} & \textbf{0.964} & \textbf{0.073} & 0.42 & \textbf{28.6} & \textbf{26.39} & \textbf{0.949} & \textbf{0.089} \\
ENeRF+Ours ($4 \times 4$) & 28.21 & 0.957 & 0.088 & 0.10 & \textbf{43.6} & 26.09 & 0.942 & 0.105 \\
MVSGaussian+Ours & 28.40 & 0.962 & \underline{0.076} & 1 & 23.4 & \underline{26.16} & 0.946 & \underline{0.093} \\
\hline
\end{tabular}
\caption{Quantitative results of generalization on the DTU testset \cite{aanaes2016large} ($512 \times 640$). The best result is highlighted in bold, while the second-best is underlined.}
\label{tab:dtu_generalization}
\end{table*}

\begin{table*}[t]
\small
\centering
\begin{tabular}{ccccccccccc}
\hline
\multirow{2}{*}{Methods} & \multirow{2}{*}{Settings} & \multicolumn{3}{c}{Real Forward-facing} & \multicolumn{3}{c}{NeRF Synthetic} \\
\cmidrule(r){3-5} \cmidrule(r){6-8}
 & & PSNR $\uparrow$ & SSIM $\uparrow$ & LPIPS $\downarrow$ & PSNR $\uparrow$ & SSIM $\uparrow$ & LPIPS $\downarrow$ \\
\hline
PixelNeRF \cite{yu2021pixelnerf} & \multirow{11}{*}{3-view} & 11.24 & 0.486 & 0.671 & 7.39 & 0.658 & 0.411 \\
IBRNet \cite{wang2021ibrnet} & & 21.79 & 0.786 & 0.279 & 22.44 & 0.874 & 0.195 \\
MVSNeRF \cite{chen2021mvsnerf} & & 21.93 & 0.795 & 0.252 & 23.62 & 0.897 & 0.176 \\
ENeRF \cite{lin2022efficient} & & 23.63 & 0.843 & 0.182 & 26.17 & 0.943 & 0.085 \\
MatchNeRF \cite{chen2023matchnerf} & & 22.43 & 0.805 & 0.244 & 23.20 & 0.897 & 0.164 \\
GNT \cite{wang2023attention} & & 22.98 & 0.761 & 0.221 & 25.80 & 0.905 & 0.104 \\
CG-NeRF \cite{nguyen2023cascaded} & & 23.93 & 0.820 & 0.210 & 25.01 & 0.900 & 0.190 \\
MuRF \cite{xu2024murf} & & 23.70 & \textbf{0.860} & 0.181 & 24.37 & 0.885 & 0.117 \\
MVSGaussian \cite{liu2024mvsgaussian} & & 24.07 & 0.857 & \underline{0.164} & 26.46 & \textbf{0.948} & \underline{0.071} \\
ENeRF+Ours ($2 \times 2$) & & \textbf{24.33} & \textbf{0.860} & \textbf{0.162} & \textbf{26.49} & \textbf{0.948} & 0.075 \\
ENeRF+Ours ($4 \times 4$) & & 23.84 & 0.851 & 0.174 & 26.00 & 0.943 & 0.083 \\
MVSGaussian+Ours & & \underline{24.15} & \underline{0.858} & 0.165 & \underline{26.48} & \underline{0.947} & \textbf{0.070} \\
\hline
MVSNeRF \cite{chen2021mvsnerf} & \multirow{8}{*}{2-view} & 20.22 & 0.763 & 0.287 & 20.56 & 0.856 & 0.243 \\
ENeRF \cite{lin2022efficient} & & 22.78 & 0.821 & 0.191 & 24.83 & 0.931 & 0.117 \\
MatchNeRF \cite{chen2023matchnerf} & & 20.59 & 0.775 & 0.276 & 20.57 & 0.864 & 0.200 \\
GNT \cite{wang2023attention} & & 20.91 & 0.683 & 0.293 & 23.47 & 0.877 & 0.151 \\
MuRF \cite{xu2024murf} & & 22.55 & 0.820 & 0.218 & 22.96 & 0.866 & 0.137 \\
MVSGaussian \cite{liu2024mvsgaussian} & & \underline{23.11} & \textbf{0.834} & \textbf{0.175} & \underline{25.06} & \textbf{0.937} & \textbf{0.079} \\
ENeRF+Ours ($2 \times 2$) & & 23.06 & 0.824 & \underline{0.186} & 25.01 & \textbf{0.937} & \underline{0.087} \\
ENeRF+Ours ($4 \times 4$) & & 23.06 & \underline{0.826} & 0.187 & 24.65 & \underline{0.932} & 0.098 \\
MVSGaussian+Ours & & \textbf{23.13} & \textbf{0.834} & \textbf{0.175} & \textbf{25.10} & \textbf{0.937} & \textbf{0.079} \\
\hline
\end{tabular}
\caption{Cross-dataset evaluation results on Real Forward-facing \cite{mildenhall2019local} ($640 \times 960$) and NeRF Synthetic \cite{mildenhall2020nerf} ($800 \times 640$) datasets. The best result is highlighted in bold, while the second-best is underlined.}
\label{tab:other_generalization}
\end{table*}

\paragraph{Datasets.} Following MVSNeRF \cite{chen2021mvsnerf, lin2022efficient, liu2024mvsgaussian},we train the generalizable model on the DTU training set \cite{aanaes2016large} and evaluate it on the DTU test set. We also perform cross-dataset evaluation on the Real Forward-facing \cite{mildenhall2019local} and NeRF Synthetic \cite{mildenhall2020nerf} datasets using the trained model without fine-tuning. These datasets introduce significant variations in view distribution and scene content compared to the DTU dataset. The quality of the synthesized novel views is measured using the PSNR, SSIM \cite{wang2004image}, and LPIPS \cite{zhang2018unreasonable} metrics.

\paragraph{Baselines.} We compare our method against several state-of-the-art generalizable NeRF methods \cite{yu2021pixelnerf, wang2021ibrnet, chen2021mvsnerf, lin2022efficient, chen2023matchnerf, nguyen2023cascaded, tanay2024global, xu2024murf}, as well as the recent generalizable 3D-GS method \cite{liu2024mvsgaussian}. We follow the same experimental settings as \cite{chen2021mvsnerf, lin2022efficient, liu2024mvsgaussian} and borrow some results reported in the respective papers. For additional baselines, such as \cite{wang2023attention, xu2024murf}, we evaluate their performance using publicly released models.

\paragraph{Implementation Details.} To verify the effectiveness of the proposed depth-guided bundle sampling, we integrate it with two representative methods, ENeRF \cite{lin2022efficient} and MVSGaussian \cite{liu2024mvsgaussian}, keeping hyperparameters consistent with their original configurations. Both models are trained with the loss function from \cite{liu2024mvsgaussian}. We observed that direct training can sometimes result in instability during the initial stages. To address this, we employ a pre-training phase that uniformly samples $N_{\text{max}}$ spheres within each bundle’s depth range for 100 epochs. For the depth-guided adaptive sampling phase, we set $N_{\text{max}} = 6$ and $\delta_s$ to 1/64 of the scene depth range, as defined in \cref{eq:num-sample}. For ENeRF+Ours, we evaluate different bundle sizes to assess the effectiveness of bundle sampling, with $2 \times 2$ and $4 \times 4$ indicating the bundle size. All models are trained and tested on an RTX A6000 GPU using the Adam optimizer \cite{kingma2014adam}.

\subsection{Results}
\begin{figure*}[t]
  \centering
   \includegraphics[width=0.95\linewidth]{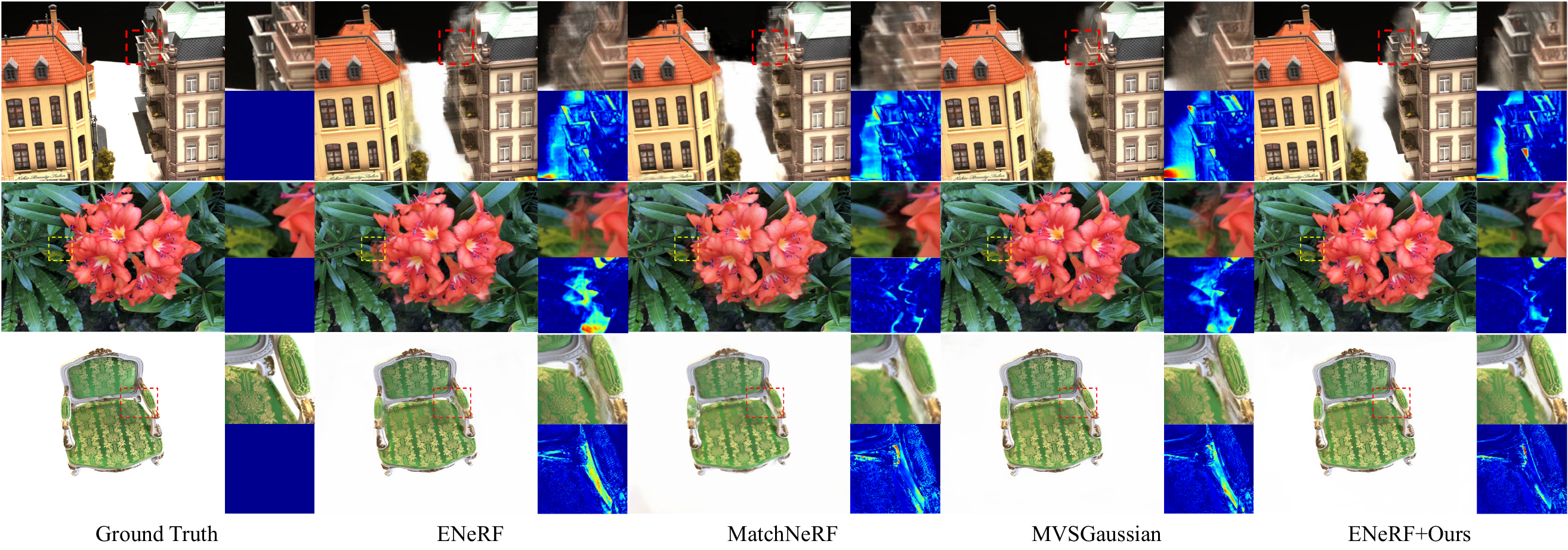}
   \caption{Qualitative comparison of ENeRF+Ours ($2 \times 2$) with state-of-the-art methods \cite{lin2022efficient, chen2023matchnerf, liu2024mvsgaussian} under 3-view setting. Each image triplet includes: the reconstructed image on the left, a zoomed-in view on the upper right, and the error map on the lower right.}
   \label{fig:vis}
\end{figure*}

We report quantitative results on the DTU test set in \cref{tab:dtu_generalization}. Additional results on two other datasets are provided in \cref{tab:other_generalization}. Notably, ENeRF \cite{lin2022efficient} and MVSGaussian \cite{liu2024mvsgaussian} achieve promising speeds by sampling only 1-2 points per ray, while other methods experience slower inference times due to the large number of samples required for accurate reconstruction. We can see that ENeRF+Ours ($2 \times 2$) improves PSNR by 1.27 dB and increases FPS by 47\% compared to original ENeRF on DTU. Additionally, ENeRF+Ours ($4 \times 4$) demonstrates better performance with over twice the speed of the original ENeRF. Performance is not significantly impacted by bundle size. For MVSGaussian+Ours, improvements are limited due to fewer sampling points and dependency on additional post-processing network layers; further performance gains may require a more integrated approach. Thanks to our efficient depth-guided bundle sampling strategy, ENeRF+Ours achieves competitive performance compared to state-of-the-art approaches, while also maintaining high inference speeds. The ability to adjust bundle size provides flexibility, allowing our method to balance speed and quality across different practical applications. Qualitative results in \cref{fig:vis} show that ENeRF+Ours ($2 \times 2$) generates high-quality views with enhanced scene detail and fewer artifacts.

\subsection{Ablations and Analysis}
We conduct a series of ablation studies to evaluate the effectiveness of our design. For efficiency, we limited our experiments to a 3-view configuration on the DTU dataset, employing bundle sizes of $2 \times 2$ and $4 \times 4$ on ENeRF \cite{lin2022efficient}.

\begin{table}[t]
\small
\centering
\begin{tabular}{ccccccc}
\hline
\multicolumn{2}{c}{Sphere} & \multirow{2}{*}{\ding{55}} & \multirow{2}{*}{\checkmark} & \multirow{2}{*}{\checkmark} & \multirow{2}{*}{\checkmark} & \multirow{2}{*}{\checkmark} \\
\multicolumn{2}{c}{Sampling} &  &  &  &  & \\
\hline
\multicolumn{2}{c}{Adaptive} & \multirow{2}{*}{\checkmark} & \multirow{2}{*}{\ding{55}} & \multirow{2}{*}{\checkmark} & \multirow{2}{*}{\checkmark} & \multirow{2}{*}{\checkmark} \\
\multicolumn{2}{c}{Sampling} &  &  &  &  & \\
\hline
\multicolumn{2}{c}{Joint Bundle} & \multirow{2}{*}{\checkmark} & \multirow{2}{*}{\checkmark} & \multirow{2}{*}{\checkmark} & \multirow{2}{*}{\ding{55}} & \multirow{2}{*}{\checkmark} \\
\multicolumn{2}{c}{Representation} &  &  &  &  & \\
\hline
\multicolumn{2}{c}{Ray-Specific} & \multirow{2}{*}{\checkmark} & \multirow{2}{*}{\checkmark} & \multirow{2}{*}{\ding{55}} & \multirow{2}{*}{\checkmark} & \multirow{2}{*}{\checkmark} \\
\multicolumn{2}{c}{Representation} &  &  &  &  & \\
\hline
\multirow{4}{*}{$2 \times 2$} & PSNR $\uparrow$ & 27.66 & 28.85 & 28.47 & 27.83 & \textbf{28.86} \\
 & SSIM $\uparrow$ & 0.949 & \textbf{0.964} & 0.958 & 0.959 & \textbf{0.964} \\
 & LPIPS $\downarrow$ & 0.090 & 0.076 & 0.078 & 0.933 & \textbf{0.073} \\
 & FPS $\uparrow$ & 29.2 & 17.0 & 29.4 & \textbf{33.74} & 28.6 \\
\hline
\multirow{4}{*}{$4 \times 4$} & PSNR $\uparrow$ & 25.93 & 28.06 & 26.74 & 27.42 & \textbf{28.21} \\
 & SSIM $\uparrow$ & 0.940 & 0.954 & 0.928 & 0.955 & \textbf{0.957} \\
 & LPIPS $\downarrow$ & 0.105 & 0.095 & 0.112 & 0.099 & \textbf{0.088} \\
 & FPS $\uparrow$ & 44.2 & 29.1 & 45.3 & \textbf{49.6} & 43.6 \\
\hline
\end{tabular}
\caption{Ablations. The best result is highlighted in bold.}
\label{tab:ablations}
\end{table}

\paragraph{Sampling Strategy.} A unique aspect of our approach is the sphere-based bundle sampling (see \cref{fig:bundle_sampling}), where we sample features from the mipmap of source view feature maps according to the area occupied by each sampled sphere. As shown in \cref{tab:ablations}, substituting our sphere-based sampling with a pixel-aligned feature extraction at sphere centers results in a PSNR decrease of 1.2 dB.

\begin{figure}[t]
  \centering
  \subfloat[Target View]{
    \includegraphics[width=0.47\columnwidth]{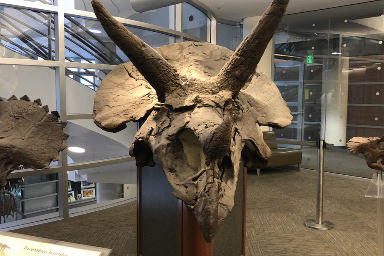}
  }
  \subfloat[Sample Distribution]{
    \includegraphics[width=0.47\columnwidth]{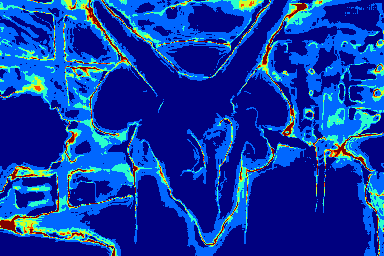}
  }
  \caption{Visualization of sample allocation in depth-guided adaptive sampling.}
  \label{fig:adaptive_sampling}
\end{figure}

\paragraph{Depth-Guided Strategy.} The proposed depth-guided adaptive sampling strategy enhances efficiency by not only skipping empty regions but also dynamically adjusting the number of samples according to depth confidence. As illustrated in \cref{fig:adaptive_sampling}, this approach allocates more samples at object boundaries and areas with depth discontinuities, where detail is essential. As presented in \cref{tab:ablations}, this method achieves a 50\% increase in FPS while maintaining comparable image quality.

\paragraph{Encoding Strategy.} The joint bundle representation, containing pre-filtered features for the sampled spheres, is inherently smooth and lacks high-frequency details. To enhance detail preservation, we integrate a ray-specific representation that provides pixel-aligned colors for individual ray samples within each sphere. As illustrated in \cref{tab:ablations}, adding this ray-specific representation boosts PSNR by 0.39 dB for a $2 \times 2$ bundle configuration and 1.47 dB for a $4 \times 4$ bundle configuration, with slight impact on FPS.

\section{Conclusion}
We proposed a novel bundle sampling strategy for generalizable novel view synthesis, inspired by plenoptic sampling theory to exploit the piece-wise smoothness of natural scenes. By grouping adjacent rays into bundles and sampling collectively, our approach reduces sampling redundancy while preserving rendering quality. Additionally, our depth-guided adaptive sampling method further optimizes efficiency by dynamically adjusting the number of samples based on depth confidence, concentrating samples at complex regions such as object boundaries and depth discontinuities. Applied to existing methods, our method achieved significant improvements in speed and fidelity. Experimental results on multiple datasets show that our method provides competitive quality with a substantial speed advantage over existing generalizable radiance field models. The flexibility to adjust bundle size also enables trade-offs between speed and quality, broadening its practical applicability to high-resolution, real-time applications.
{
    \small
    \bibliographystyle{ieeenat_fullname}
    \bibliography{main.bib}
}
\clearpage
\maketitleagain{Supplementary Material}

\section{Implementation and Network Details}
\paragraph{Implementation Details.} Following ENeRF \cite{lin2022efficient} and MVSGaussian \cite{liu2024mvsgaussian}, we partition the DTU dataset \cite{aanaes2016large} into 88 training scenes and 16 test scenes. The generalizable model is trained on the 88 training scenes and evaluated on the 16 test scenes using an RTX A6000 GPU. Training is performed with the Adam optimizer \cite{kingma2014adam}, starting with an initial learning rate of $5 \times 10^{-4}$. The batch size is set to 4, and the learning rate is halved every 50k iterations. During training, source views are selected with probabilities of 0.1, 0.8, and 0.1 for 2, 3, and 4 input views, respectively. Evaluation is conducted following the criteria outlined in prior works such as ENeRF \cite{lin2022efficient} and MVSGaussian \cite{liu2024mvsgaussian}. Specifically, for the DTU test set, segmentation masks are employed to evaluate performance, defined based on the availability of ground-truth depth at each pixel. For the Real Forward-facing dataset \cite{mildenhall2019local}, where marginal regions are typically occluded in the input images, evaluations are restricted to the central 80\% of the image area. The image resolutions used for evaluation are $512 \times 640$ for the DTU dataset \cite{aanaes2016large}, $640 \times 960$ for the Real Forward-facing dataset \cite{mildenhall2019local}, and $800 \times 800$ for the NeRF Synthetic dataset \cite{mildenhall2020nerf}. 

\begin{figure}[t]
  \centering
  \subfloat[Residual Dense Network for Neural Renderer]{
    \includegraphics[width=0.95\columnwidth]{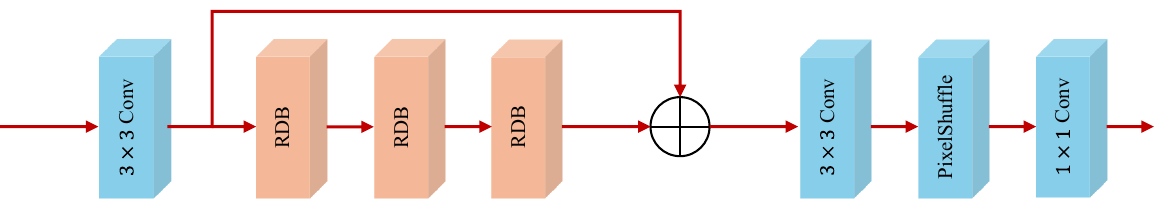}
  }
  \quad
  \subfloat[Residual Dense Block]{
    \includegraphics[width=0.95\columnwidth]{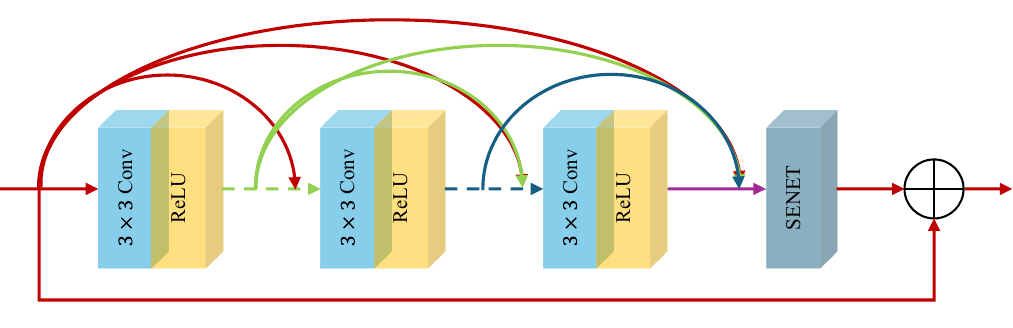}
  }
  \caption{Neural renderer is a (a) residual dense network to decode joint bundle representation into colors, which consists of 3 (b) residual dense blocks.}
  \label{fig:renderer}
\end{figure}

\paragraph{Network Details.} The mipmap query process is a well-optimized technique in the rendering community for efficient texture sampling. To leverage this, we employ the nvdiffrast library \cite{laine2020modular} to implement our bundle sampling with high efficiency. The radiance field prediction network described in Eq. (7) in Section 4.3 of the main text is adapted from ENeRF \cite{lin2022efficient}, with modifications to the feature dimensions of the samples. The neural renderer comprises three residual dense blocks \cite{zhang2018residual} integrated with SENet \cite{hu2018squeeze}, as illustrated in \cref{fig:renderer}. Each block employs a hidden dimension of 64 and a growth rate of 32. The ray-specific representation is unfolded using a pixel shuffle operation. During training, the final target view is synthesized by fusing two intermediate views, calculated as: $\boldsymbol{\hat{I}}_{\text{tar}} = \boldsymbol{\hat{I}}_c + \boldsymbol{\hat{I}}_f$. However, for the Real Forward-facing \cite{mildenhall2019local} and NeRF Synthetic \cite{mildenhall2020nerf} datasets—domains that differ significantly from the DTU dataset—adjusting the weights of the intermediate views improves performance. Specifically, during evaluation on these datasets, the final view is generated as: $\boldsymbol{\hat{I}}_{\text{tar}} = 0.5 \boldsymbol{\hat{I}}_c + 1.5 \boldsymbol{\hat{I}}_f$. This weighted fusion strategy effectively adapts the model to cross-domain scenarios, enhancing rendering performance.

\section{Additional Ablation Experiments}
To demonstrate the effectiveness of the proposed bundle sampling strategy, we apply a $1 \times 1$ bundle size to ENeRF, which corresponds to casting each cone through a single pixel. This setup disables bundle sampling while preserving all other components of the method. The results on the DTU test set are presented in \cref{tab:bundle_size_dtu}, while cross-dataset results on the Real Forward-facing and NeRF Synthetic datasets are shown in \cref{tab:bundle_size_other}.

\begin{table*}[t]
\small
\centering
\begin{tabular}{ccccccccc}
\hline
\multirow{2}{*}{Methods} & \multicolumn{5}{c}{3-view} & \multicolumn{3}{c}{2-view} \\
\cmidrule(r){2-6} \cmidrule(r){7-9}
 & PSNR $\uparrow$ & SSIM $\uparrow$ & LPIPS $\downarrow$ & Avg. samples per ray & FPS $\uparrow$ & PSNR $\uparrow$ & SSIM $\uparrow$ & LPIPS $\downarrow$ \\
\hline
ENeRF+Ours ($1 \times 1$) & \underline{28.75} & \textbf{0.966} & \textbf{0.069} & 1.63 & 15.3 & \underline{26.37} & \textbf{0.951} & \textbf{0.088} \\
ENeRF+Ours ($2 \times 2$) & \textbf{28.86} & \underline{0.964} & \underline{0.073} & 0.42 & \underline{28.6} & \textbf{26.39} & \underline{0.949} & \underline{0.089} \\
ENeRF+Ours ($4 \times 4$) & 28.21 & 0.957 & 0.088 & 0.10 & \textbf{43.6} & 26.09 & 0.942 & 0.105 \\
\hline
\end{tabular}
\caption{Ablation on bundle size on the DTU testset \cite{aanaes2016large}. The best result is highlighted in bold, while the second-best is underlined.}
\label{tab:bundle_size_dtu}
\end{table*}

\begin{table*}[t]
\centering
\begin{tabular}{ccccccccccc}
\hline
\multirow{2}{*}{Methods} & \multirow{2}{*}{Settings} & \multicolumn{3}{c}{Real Forward-facing} & \multicolumn{3}{c}{NeRF Synthetic} \\
\cmidrule(r){3-5} \cmidrule(r){6-8}
 & & PSNR $\uparrow$ & SSIM $\uparrow$ & LPIPS $\downarrow$ & PSNR $\uparrow$ & SSIM $\uparrow$ & LPIPS $\downarrow$ \\
\hline
ENeRF+Ours ($1 \times 1$) & \multirow{3}{*}{3-view} & \underline{24.20} & \textbf{0.860} & \underline{0.164} & \textbf{26.62} & \textbf{0.948} & \underline{0.076} \\
ENeRF+Ours ($2 \times 2$) & & \textbf{24.33} & \textbf{0.860} & \textbf{0.162} & \underline{26.48} & \textbf{0.948} & \textbf{0.075} \\
ENeRF+Ours ($4 \times 4$) & & 23.84 & \underline{0.850} & 0.174 & 26.12 & \underline{0.943} & 0.104 \\
\hline
ENeRF+Ours ($1 \times 1$) & \multirow{3}{*}{2-view} & \textbf{23.14} & \underline{0.825} & \underline{0.187} & \underline{24.98} & \underline{0.936} & \underline{0.097} \\
ENeRF+Ours ($2 \times 2$) & & \underline{23.06} & 0.824 & \textbf{0.186} & \textbf{25.01} & \textbf{0.937} & \textbf{0.087} \\
ENeRF+Ours ($4 \times 4$) & & \underline{23.06} & \textbf{0.826} & \underline{0.187} & 24.65 & 0.932 & 0.098 \\
\hline
\end{tabular}
\caption{Ablation on bundle size on Real Forward-facing \cite{mildenhall2019local} and NeRF Synthetic \cite{mildenhall2020nerf}. The best result is highlighted in bold, while the second-best is underlined.}
\label{tab:bundle_size_other}
\end{table*}

As observed, the quality difference between $1 \times 1$ and $2 \times 2$ bundle sizes is quite small, demonstrating that our bundle sampling strategy effectively reduces the number of samples without compromising rendering quality. Notably, in the natural scenes of Real Forward-facing dataset, a $4 \times 4$ bundle size achieves results comparable to those of smaller bundle sizes. This further validates the efficiency of our method, particularly for scenes characterized by piece-wise smoothness.

\section{Per-Scene Optimization Results}
For per-scene optimization, we follow the setting of ENeRF \cite{lin2022efficient}, selecting 3 and 4 source views with probabilities of 0.4 and 0.6, respectively. The batch size is set to 1. For efficiency, the bundle size is fixed at $2 \times 2$ for ENeRF+Ours. The generalizable model is fine-tuned on each scene for 10 epochs (less than 1 hour). For evaluation, 4 input views are used.

\begin{table*}[!ht]
\small
\centering
\begin{tabular}{ccccccccccc}
\hline
\multirow{2}{*}{Methods} & \multicolumn{3}{c}{DTU} & \multicolumn{3}{c}{Real Forward-facing} & \multicolumn{3}{c}{NeRF Synthetic} \\
\cmidrule(r){2-4} \cmidrule(r){5-7} \cmidrule(r){8-10}
 & PSNR $\uparrow$ & SSIM $\uparrow$ & LPIPS $\downarrow$ & PSNR $\uparrow$ & SSIM $\uparrow$ & LPIPS $\downarrow$ & PSNR $\uparrow$ & SSIM $\uparrow$ & LPIPS $\downarrow$ \\
\hline
$\text{NeRF}_{10.2h}$ \cite{mildenhall2020nerf} & 27.01 & 0.902 & 0.263 & 25.97 & 0.870 & 0.236 & \textbf{30.63} & \textbf{0.962} & 0.093 \\
$\text{IBRNet}_{ft-1.0h}$ \cite{wang2021ibrnet} & \textbf{31.35} & \underline{0.956} & 0.131 & 24.88 & 0.861 & 0.189 & 25.62 & 0.939 & 0.111 \\
$\text{MVSNeRF}_{ft-15min}$ \cite{chen2021mvsnerf} & 28.51 & 0.933 & 0.179 & 25.45 & 0.877 & 0.192 &  27.07 & 0.931 & 0.168 \\
$\text{ENeRF}_{ft-1.0h}$ \cite{lin2022efficient} & 28.87 & 0.957 & 0.090 & 24.89 & 0.865 & 0.159 & 27.57 & 0.954 & 0.063 \\
$\text{MVSGaussian}_{ft-1.0h}$ \cite{liu2024mvsgaussian} & - & - & - & 25.92 & 0.891 & 0.135 & 27.87 & 0.956 & \underline{0.061} \\
$\text{ENeRF+Ours}_{ft-15min}$ & 29.57 & \textbf{0.962} & \underline{0.079} & \underline{26.29} & \underline{0.894} & \underline{0.125} & 28.07 & 0.957 & 0.062 \\
$\text{ENeRF+Ours}_{ft-1.0h}$ & \underline{29.78} & \textbf{0.962} & \textbf{0.077} & \textbf{26.75} & \textbf{0.904} & \textbf{0.113} & \underline{28.39} & \underline{0.958} & \textbf{0.060} \\
\hline
\end{tabular}
\caption{Per-scene optimization results on DTU \cite{aanaes2016large}, Real Forward-facing \cite{mildenhall2019local} and NeRF Synthetic \cite{mildenhall2020nerf} datasets. The best result is highlighted in bold, while the second-best is underlined.}
\label{tab:per-scene-optim}
\end{table*}

The per-scene optimization results on the DTU, Real Forward-facing and NeRF Synthetic datasets are presented in \cref{tab:per-scene-optim}. The results demonstrate that our method outperforms other approaches on most scenes, particularly in natural scenes of the Real Forward-facing dataset. On the NeRF Synthetic dataset, our method achieves second-best performance in terms of PSNR and SSIM, trailing only NeRF \cite{mildenhall2020nerf}. However, our method significantly outperforms NeRF in LPIPS, indicating superior perceptual quality. Furthermore, our approach achieves this performance while requiring only 1/10 of NeRF's training time, highlighting its efficiency.

\section{Per-Scene Breakdown Results}
The per-scene breakdown results are presented in \cref{tab:per-scene-dtu} and \cref{tab:per-scene-other-dtu} for the DTU dataset, \cref{tab:per-scene-llff} for the Real Forward-facing dataset, and \cref{tab:per-scene-nerf} for the NeRF Synthetic dataset. For efficiency, the bundle size is fixed at $2 \times 2$ for ENeRF+Ours. These detailed results are consistent with the averaged results reported in the main text.

\begin{table*}[t]
\centering
\begin{tabular}{c|ccccc}
\hline
Scan & \#1 & \#8 & \#21 & \#103 & \#114 \\
\hline
Metric & \multicolumn{5}{c}{PSNR} \\
\hline
PixelNeRF \cite{yu2021pixelnerf} & 21.64 & 23.70 & 16.04 & 16.76 & 18.40 \\
IBRNet \cite{wang2021ibrnet} & 25.97 & 27.45 & 20.94 & 27.91 & 27.91 \\
MVSNeRF \cite{chen2021mvsnerf} & 26.96 & 27.43 & 21.55 & 29.25 & 27.99 \\
ENeRF \cite{lin2022efficient} & 28.85 & 29.05 & 22.53 & 30.51 & 28.86 \\
MatchNeRF \cite{chen2023matchnerf} & 27.69 & 27.76 & 22.75 & 29.35 & 28.16 \\
GNT \cite{wang2023attention} & 27.25 & 28.12 & 21.67 & 28.45 & 28.01 \\
MVSGaussian \cite{liu2024mvsgaussian} & \textbf{29.67} & \underline{29.65} & \underline{23.24} & \underline{30.60} & \underline{29.26} \\
ENeRF+Ours & \underline{29.62} & \textbf{30.42} & \textbf{23.40} & \textbf{32.28} & \textbf{30.11} \\
\hline
$\text{NeRF}_{10.2h}$ \cite{mildenhall2020nerf} & 26.62 & 28.33 & 23.24 & 30.40 & 26.47 \\
$\text{IBRNet}_{ft-1.0h}$ \cite{wang2021ibrnet} & \textbf{31.00} & \textbf{32.46} & \textbf{27.88} & \textbf{34.40} & \textbf{31.00} \\
$\text{MVSNeRF}_{ft-15min}$ \cite{chen2021mvsnerf} & 28.05 & 28.88 & \underline{24.87} & 32.23 & 28.47 \\
$\text{ENeRF}_{ft-1.0h}$ \cite{lin2022efficient} & 30.10 & 30.50 & 22.46 & 31.42 & 29.87 \\
$\text{ENeRF+Ours}_{ft-15min}$ & 30.11 & 31.09 & 23.51 & 32.79 & 30.34 \\
$\text{ENeRF+Ours}_{ft-1.0h}$ & \underline{30.80} & \underline{31.33} & 23.59 & \underline{32.85} & \underline{30.35} \\
\hline
Metric & \multicolumn{5}{c}{SSIM} \\
\hline
PixelNeRF \cite{yu2021pixelnerf} & 0.827 & 0.829 & 0.691 & 0.836 & 0.763 \\
IBRNet \cite{wang2021ibrnet} & 0.918 & 0.903 & 0.873 & 0.950 & 0.943 \\
MVSNeRF \cite{chen2021mvsnerf} & 0.937 & 0.922 & 0.890 & 0.962 & 0.949 \\
ENeRF \cite{lin2022efficient} & 0.958 & 0.955 & 0.916 & 0.968 & 0.961 \\
MatchNeRF \cite{chen2023matchnerf} & 0.936 & 0.918 & 0.901 & 0.961 & 0.948 \\
GNT \cite{wang2023attention} & 0.922 & 0.931 & 0.881 & 0.942 & 0.960 \\
MVSGaussian \cite{liu2024mvsgaussian} & \textbf{0.966} & \textbf{0.961} & \underline{0.930} & \underline{0.970} & \underline{0.963} \\
ENeRF+Ours & \underline{0.965} & \underline{0.960} & \textbf{0.939} & \textbf{0.974} & \textbf{0.966} \\
\hline
$\text{NeRF}_{10.2h}$ \cite{mildenhall2020nerf} & 0.902 & 0.876 & 0.874 & 0.944 & 0.913 \\
$\text{IBRNet}_{ft-1.0h}$ \cite{wang2021ibrnet} & 0.955 & 0.945 & \textbf{0.947} & 0.968 & 0.964 \\
$\text{MVSNeRF}_{ft-15min}$ \cite{chen2021mvsnerf} & 0.934 & 0.900 & 0.922 & 0.964 & 0.945 \\
$\text{ENeRF}_{ft-1.0h}$ \cite{lin2022efficient} & 0.966 & 0.959 & 0.924 & \underline{0.971} & \underline{0.965} \\
$\text{ENeRF+Ours}_{ft-15min}$ & \underline{0.967} & \underline{0.961} & 0.939 & \textbf{0.975} & \textbf{0.967} \\
$\text{ENeRF+Ours}_{ft-1.0h}$ & \textbf{0.968} & \textbf{0.962} & \underline{0.940} & \textbf{0.975} & \textbf{0.967} \\
\hline
Metric & \multicolumn{5}{c}{LPIPS} \\
\hline
PixelNeRF \cite{yu2021pixelnerf} & 0.373 & 0.384 & 0.407 & 0.376 & 0.372 \\
IBRNet \cite{wang2021ibrnet} & 0.190 & 0.252 & 0.179 & 0.195 & 0.136 \\
MVSNeRF \cite{chen2021mvsnerf} & 0.155 & 0.220 & 0.166 & 0.165 & 0.135 \\
ENeRF \cite{lin2022efficient} & 0.086 & 0.119 & 0.107 & 0.107 & 0.076 \\
MatchNeRF \cite{chen2023matchnerf} & 0.157 & 0.227 & 0.149 & 0.179 & 0.132 \\
GNT \cite{wang2023attention} & 0.143 & 0.210 & 0.171 & 0.149 & 0.139 \\
MVSGaussian \cite{liu2024mvsgaussian} & \textbf{0.069} & \textbf{0.102} & 0.088 & 0.098 & \textbf{0.070} \\
ENeRF+Ours & \underline{0.070} & \underline{0.105} & \textbf{0.076} & \textbf{0.091} & \underline{0.071} \\
\hline
$\text{NeRF}_{10.2h}$ \cite{mildenhall2020nerf} & 0.265 & 0.321 & 0.246 & 0.256 & 0.225 \\
$\text{IBRNet}_{ft-1.0h}$ \cite{wang2021ibrnet} & 0.129 & 0.170 & 0.104 & 0.156 & 0.099 \\
$\text{MVSNeRF}_{ft-15min}$ \cite{chen2021mvsnerf} & 0.171 & 0.261 & 0.142 & 0.170 & 0.153 \\
$\text{ENeRF}_{ft-1.0h}$ \cite{lin2022efficient} & 0.071 & 0.106 & \underline{0.097} & 0.102 & 0.074 \\
$\text{ENeRF+Ours}_{ft-15min}$ & \underline{0.066} & \underline{0.097} & \textbf{0.074} & \underline{0.086} & \underline{0.066} \\
$\text{ENeRF+Ours}_{ft-1.0h}$ & \textbf{0.064} & \textbf{0.096} & \textbf{0.074} & \textbf{0.084} & \textbf{0.065} \\
\hline
\end{tabular}
\caption{Quantitative results of five sample scenes on the DTU \cite{aanaes2016large} test set. The best result is highlighted in bold, while the second-best is underlined.}
\label{tab:per-scene-dtu}
\end{table*}

\begin{table*}[t]
\centering
\begin{tabular}{c|ccccccccccc}
\hline
Scan & \#30 & \#31 & \#34 & \#38 & \#40 & \#41 & \#45 & \#55 & \#63 & \#82 & \#110 \\
\hline
Metric & \multicolumn{11}{c}{PSNR} \\
\hline
ENeRF \cite{lin2022efficient} & 29.20 & 25.13 & 26.77 & 28.61 & 25.67 & 29.51 & 24.83 & 30.26 & 27.22 & 26.83 & 27.97 \\
MatchNeRF \cite{chen2023matchnerf} & 29.16 & 24.26 & 25.66 & 27.52 & 25.16 & 28.27 & 23.94 & 26.64 & \textbf{29.40} & 27.65 & 27.15 \\
GNT \cite{wang2023attention} & 27.13 & 23.54 & 25.10 & 27.67  & 24.48 & 28.10 & 24.54 & 28.86 & 26.36 & 26.09 & 26.93 \\
MVSGaussian \cite{liu2024mvsgaussian} & 30.10 & 25.94 & 26.82 & 29.27 & 26.13 & 30.33 & 24.55 & 31.40 & 28.46 & 27.82 & 28.15 \\
ENeRF+Ours & \textbf{30.89} & \textbf{26.76} & \textbf{27.81} & \textbf{30.10} & \textbf{27.16} & \textbf{30.84} & \textbf{25.95} & \textbf{31.63} & 29.39 & \textbf{29.55} & \textbf{29.47} \\
\hline
Metric & \multicolumn{11}{c}{SSIM} \\
\hline
ENeRF \cite{lin2022efficient} & 0.981 & 0.937 & 0.934 & 0.946 & 0.947 & 0.960 & 0.948 & 0.973 & 0.978 & 0.971 & 0.974 \\
MatchNeRF \cite{chen2023matchnerf} & 0.974 & 0.921 & 0.874 & 0.902 & 0.903 & 0.936 & 0.934 & 0.929 & 0.976 & 0.966 & 0.962 \\
GNT \cite{wang2023attention} & 0.954 & 0.907 & 0.880 & 0.921 & 0.893 & 0.908 & 0.918  & 0.934 & 0.938 & 0.949 & 0.930 \\
MVSGaussian \cite{liu2024mvsgaussian} & 0.983 & 0.946 & \textbf{0.947} & 0.954 & \textbf{0.957} & \textbf{0.967} & 0.954 & \textbf{0.979} & 0.980 & 0.974 & 0.976 \\
ENeRF+Ours & \textbf{0.985} & \textbf{0.953} & 0.944 & \textbf{0.955} & 0.954 & 0.966 & \textbf{0.959} & \textbf{0.979} & \textbf{0.984} & \textbf{0.979} & \textbf{0.979} \\
\hline
Metric & \multicolumn{11}{c}{LPIPS} \\
\hline
ENeRF \cite{lin2022efficient} & 0.052 & 0.108 & 0.117 & 0.118 & 0.120 & 0.091 & 0.077 & 0.069 & 0.048 & 0.066 & 0.069 \\
MatchNeRF \cite{chen2023matchnerf} & 0.085 & 0.169 & 0.234 & 0.220 & 0.216 & 0.174 & 0.127 & 0.164 & 0.077 & 0.093 & 0.141 \\
GNT \cite{wang2023attention} & 0.110 & 0.172 & 0.201 & 0.231 & 0.116 & 0.168 & 0.134 & 0.155 & 0.127 & 0.138 & 0.127 \\
MVSGaussian \cite{liu2024mvsgaussian} & 0.048 & 0.093 & \textbf{0.097} & 0.098 & \textbf{0.101} & \textbf{0.075} & 0.067 & 0.055 & 0.041 & 0.057 & \textbf{0.057} \\
ENeRF+Ours & \textbf{0.045} & \textbf{0.083} & 0.103 & \textbf{0.095} & \textbf{0.101} & \textbf{0.075} & \textbf{0.057} & \textbf{0.054} & \textbf{0.035} & \textbf{0.048} & 0.060 \\
\hline
\end{tabular}
\caption{Quantitative results of other eleven scenes on the DTU \cite{aanaes2016large} test set. The best result is highlighted in bold, while the second-best is underlined.}
\label{tab:per-scene-other-dtu}
\end{table*}

\begin{table*}[t]
\centering
\begin{tabular}{c|cccccccc}
\hline
Scene & Fern & Flower & Fortress & Horns & Leaves & Orchids & Room & Trex \\
\hline
Metric & \multicolumn{8}{c}{PSNR} \\
\hline
PixelNeRF \cite{yu2021pixelnerf} & 12.40 & 10.00 & 14.07 & 11.07 & 9.85 & 9.62 & 11.75 & 10.55 \\
IBRNet \cite{wang2021ibrnet} & 20.83 & 22.38 & 27.67 & 22.06 & 18.75 & 15.29 & 27.26 & 20.06 \\
MVSNeRF \cite{chen2021mvsnerf} & 21.15 & 24.74 & 26.03 & 23.57 & 17.51 & 17.85 & 26.95 & \underline{23.20} \\
ENeRF \cite{lin2022efficient} & 21.92 & 24.28 & 30.43 & 24.49 & 19.01 & 17.94 & 29.75 & 21.21 \\
MatchNeRF \cite{chen2023matchnerf} & 20.98 & 23.97 & 27.44 & 23.14 & 18.62 & \underline{18.07} & 26.77 & 20.47 \\
GNT \cite{wang2023attention} & 22.21 & 23.56 & 29.16 & 22.80 & 19.18 & 17.43 & 29.35 & 20.15 \\
MVSGaussian \cite{liu2024mvsgaussian} & \underline{22.45} & \textbf{25.66} & \underline{30.46} & \underline{24.70} & \underline{19.81} & 17.86 & \underline{29.86} & 21.75 \\
ENeRF+Ours & \textbf{22.55} & \underline{25.44} & \textbf{31.30} & \textbf{26.45} & \textbf{19.88} & \textbf{18.96} & \textbf{30.16} & \textbf{23.80} \\
\hline
$\text{NeRF}_{10.2h}$ \cite{mildenhall2020nerf} & \textbf{23.87} & 26.84 & 31.37 & 25.96 & 21.21 & 19.81 & \textbf{33.54} & 25.19 \\
$\text{IBRNet}_{ft-1.0h}$ \cite{wang2021ibrnet} & 22.64 & 26.55 & 30.34 & 25.01 & 22.07 & 19.01 & 31.05 & 22.34 \\
$\text{MVSNeRF}_{ft-15min}$ \cite{chen2021mvsnerf} & 23.10 & 27.23 & 30.43 & 26.35 & 21.54 & \underline{20.51} & 30.12 & 24.32 \\
$\text{ENeRF}_{ft-1.0h}$ \cite{lin2022efficient} & 22.08 & \underline{27.74} & 29.58 & 25.50 & 21.26 & 19.50 & 30.07 & 23.39 \\
$\text{ENeRF+Ours}_{ft-15min}$ & 23.65 & 27.63 & \underline{31.68} & \underline{27.62} & \underline{22.46} & 19.02 & 32.85 & \underline{25.37} \\
$\text{ENeRF+Ours}_{ft-1.0h}$ & \underline{23.81} & \textbf{28.07} & \textbf{31.69} & \textbf{27.83} & \textbf{22.72} & \textbf{20.56} & \underline{33.51} & \textbf{25.82} \\
\hline
Metric & \multicolumn{8}{c}{SSIM}                                          \\
\hline
PixelNeRF \cite{yu2021pixelnerf} & 0.531 & 0.433 & 0.674 & 0.516 & 0.268 & 0.317 & 0.691 & 0.458 \\
IBRNet \cite{wang2021ibrnet} & 0.710 & 0.854 & 0.894 & 0.840 & 0.705 & 0.571 & 0.950 & 0.768 \\
MVSNeRF \cite{chen2021mvsnerf} & 0.638 & 0.888 & 0.872 & 0.868 & 0.667 & 0.657 & 0.951 & \underline{0.868} \\
ENeRF \cite{lin2022efficient} & 0.774 & 0.893 & \underline{0.948} & 0.905 & 0.744 & 0.681 & 0.971 & 0.826 \\
MatchNeRF \cite{chen2023matchnerf} & 0.726 & 0.861 & 0.906 & 0.870 & 0.690 & 0.675 & 0.949 & 0.767 \\
GNT \cite{wang2023attention} & 0.736 & 0.791 & 0.867 & 0.820 & 0.650 & 0.538 & 0.945 & 0.744 \\
MVSGaussian \cite{liu2024mvsgaussian} & \underline{0.792} & \underline{0.908} & \underline{0.948} & \underline{0.913} & \textbf{0.784} & \underline{0.701} & \underline{0.973} & 0.841 \\
ENeRF+Ours & \textbf{0.802} & \textbf{0.911} & \textbf{0.960} & \textbf{0.934} & \underline{0.766} & \textbf{0.738} & \textbf{0.977} & \textbf{0.887} \\
\hline
$\text{NeRF}_{10.2h}$ \cite{mildenhall2020nerf} & \textbf{0.828} & 0.897 & 0.945 & 0.900 & 0.792 & 0.721 & 0.978 & 0.899 \\
$\text{IBRNet}_{ft-1.0h}$ \cite{wang2021ibrnet} & 0.774 & 0.909 & 0.937 & 0.904 & 0.843 & 0.705 & 0.972 & 0.842 \\
$\text{MVSNeRF}_{ft-15min}$ \cite{chen2021mvsnerf} & 0.795 & 0.912 & 0.943 & 0.917 & 0.826 & \underline{0.732} & 0.966 & 0.895 \\
$\text{ENeRF}_{ft-1.0h}$ \cite{lin2022efficient} & 0.770 & 0.923 & 0.940 & 0.904 & 0.827 & 0.725 & 0.965 & 0.869 \\
$\text{ENeRF+Ours}_{ft-15min}$ & 0.823 & \underline{0.928} & \textbf{0.964} & \underline{0.947} & \underline{0.868} & 0.727 & \underline{0.984} & \underline{0.912} \\
$\text{ENeRF+Ours}_{ft-1.0h}$ & \underline{0.826} & \textbf{0.932} & \underline{0.963} & \textbf{0.949} & \textbf{0.874} & \textbf{0.782} & \textbf{0.985} & \textbf{0.919} \\
\hline
Metric & \multicolumn{8}{c}{LPIPS} \\
\hline
PixelNeRF \cite{yu2021pixelnerf} & 0.650 & 0.708 & 0.608 & 0.705 & 0.695 & 0.721 & 0.611 & 0.667 \\
IBRNet \cite{wang2021ibrnet} & 0.349 & 0.224 & 0.196 & 0.285 & 0.292 & 0.413 & 0.161 & 0.314 \\
MVSNeRF \cite{chen2021mvsnerf} & 0.238 & 0.196 & 0.208 & 0.237 & 0.313 & \underline{0.274} & 0.172 & 0.184 \\
ENeRF \cite{lin2022efficient} & 0.224 & 0.164 & \underline{0.092} & 0.161 & \underline{0.216} & 0.289 & 0.120 & 0.192 \\
MatchNeRF \cite{chen2023matchnerf} & 0.285 & 0.202 & 0.169 & 0.234 & 0.277 & 0.325 & 0.167 & 0.294 \\
GNT \cite{wang2023attention} & 0.223 & 0.203 & 0.157 & 0.208 & 0.255 & 0.341 & \underline{0.103} & 0.275 \\
MVSGaussian \cite{liu2024mvsgaussian} & \textbf{0.193} & \textbf{0.133} & 0.096 & \underline{0.148} & \textbf{0.189} & 0.275 & 0.104 & \underline{0.177} \\
ENeRF+Ours & \underline{0.195} & \underline{0.136} & \textbf{0.086} & \textbf{0.125} & 0.225 & \textbf{0.254} & \textbf{0.095} & \textbf{0.148} \\
\hline
$\text{NeRF}_{10.2h}$ \cite{mildenhall2020nerf} & 0.291 & 0.176 & 0.147 & 0.247 & 0.301 & 0.321 & 0.157 & 0.245 \\
$\text{IBRNet}_{ft-1.0h}$ \cite{wang2021ibrnet} & 0.266 & 0.146 & 0.133 & 0.190 & 0.180 & 0.286 & 0.089 & 0.222 \\
$\text{MVSNeRF}_{ft-15min}$ \cite{chen2021mvsnerf} & 0.253 & 0.143 & 0.134 & 0.188 & 0.222 & 0.258 & 0.149 & 0.187 \\
$\text{ENeRF}_{ft-1.0h}$ \cite{lin2022efficient} & 0.197 & 0.121 & 0.101 & 0.155 & 0.168 & \underline{0.247} & 0.113 & 0.169 \\
$\text{ENeRF+Ours}_{ft-15min}$ & \underline{0.164} & \underline{0.099} & \underline{0.069} & \underline{0.097} & \underline{0.135} & 0.258 & \underline{0.059} & \underline{0.115} \\
$\text{ENeRF+Ours}_{ft-1.0h}$ & \textbf{0.156} & \textbf{0.093} & \textbf{0.067} & \textbf{0.093} & \textbf{0.126} & \textbf{0.201} & \textbf{0.055} & \textbf{0.109} \\
\hline
\end{tabular}
\caption{Quantitative results of other eleven scenes on the Real Forward-facing dataset \cite{mildenhall2019local}. The best result is highlighted in bold, while the second-best is underlined.}
\label{tab:per-scene-llff}
\end{table*}

\begin{table*}[t]
\small
\centering
\begin{tabular}{c|cccccccc}
\hline
Scene & Chair & Drums & Ficus & Hotdog & Lego & Materials & Mic & Ship \\
\hline
Metric & \multicolumn{8}{c}{PSNR} \\
\hline
PixelNeRF \cite{yu2021pixelnerf} & 7.18 & 8.15 & 6.61 & 6.80 & 7.74 & 7.61 & 7.71 & 7.30 \\
IBRNet \cite{wang2021ibrnet} & 24.20 & 18.63 & 21.59 & 27.70 & 22.01 & 20.91 & 22.10 & 22.36 \\
MVSNeRF \cite{chen2021mvsnerf} & 23.35 & 20.71 & 21.98 & 28.44 & 23.18 & 20.05 & 22.62 & 23.35 \\
ENeRF \cite{lin2022efficient} & 28.29 & \underline{21.71} & 23.83 & 34.20 & \underline{24.97} & 24.01 & 26.62 & 25.73 \\
MatchNeRF \cite{chen2023matchnerf} & 25.23 & 19.97 & 22.72 & 24.19 & 23.77 & 23.12 & 24.46 & 22.11 \\
GNT \cite{wang2023attention} & 27.98 & 20.27 & \textbf{26.86} & 29.34 & 23.17 & \textbf{30.75} & 23.19 & 24.86 \\
MVSGaussian \cite{liu2024mvsgaussian} & \underline{28.93} & \textbf{22.20} & 23.55 & \textbf{35.01} & \underline{24.97} & \underline{24.49} & \underline{26.80} & \underline{25.75} \\
ENeRF+Ours & \textbf{29.56} & 21.47 & \underline{23.86} & \underline{34.42} & \textbf{25.55} & 24.25 & \textbf{27.15} & \textbf{26.33} \\
\hline
$\text{NeRF}_{10.2h}$ \cite{mildenhall2020nerf} & \textbf{31.07} & \textbf{25.46} & \textbf{29.73} & 34.63 & \textbf{32.66} & \textbf{30.22} & \textbf{31.81} & \textbf{29.49} \\
$\text{IBRNet}_{ft-1.0h}$ \cite{wang2021ibrnet} & 28.18 & 21.93 & 25.01 & 31.48 & 25.34 & 24.27 & 27.29 & 21.48 \\
$\text{MVSNeRF}_{ft-15min}$ \cite{chen2021mvsnerf} & 26.80 & 22.48 & \underline{26.24} & 32.65 & 26.62 & 25.28 & \underline{29.78} & 26.73 \\
$\text{ENeRF}_{ft-1.0h}$ \cite{lin2022efficient} & 28.94 & \underline{25.33} & 24.71 & 35.63 & 25.39 & 24.98 & 29.25 & 26.36 \\
$\text{ENeRF+Ours}_{ft-15min}$ & 29.93 & 23.07 & 25.38 & \underline{36.53} & 27.01 & 26.56 & 28.81 & 27.26 \\
$\text{ENeRF+Ours}_{ft-1.0h}$ & \underline{30.32} & 23.27 & 25.50 & \textbf{36.75} & \underline{27.42} & \underline{27.08} & 29.42 & \underline{27.37} \\
\hline
Metric & \multicolumn{8}{c}{SSIM} \\
\hline
PixelNeRF \cite{yu2021pixelnerf} & 0.624 & 0.670 & 0.669 & 0.669 & 0.671 & 0.644 & 0.729 & 0.584 \\
IBRNet \cite{wang2021ibrnet} & 0.888 & 0.836 & 0.881 & 0.923 & 0.874 & 0.872 & 0.927 & 0.794 \\
MVSNeRF \cite{chen2021mvsnerf} & 0.876 & 0.886 & 0.898 & 0.962 & 0.902 & 0.893 & 0.923 & 0.886 \\
ENeRF \cite{lin2022efficient} & 0.965 & 0.918 & 0.932 & 0.981 & \underline{0.948} & 0.937 & 0.969 & 0.891 \\
MatchNeRF \cite{chen2023matchnerf} & 0.908 & 0.868 & 0.897 & 0.943 & 0.903 & 0.908 & 0.947 & 0.806 \\
GNT \cite{wang2023attention} & 0.935 & 0.891 & \textbf{0.941} & 0.940 & 0.897 & \textbf{0.974} & 0.791 & 0.874 \\
MVSGaussian \cite{liu2024mvsgaussian} & \underline{0.969} & \textbf{0.927} & 0.935 & \textbf{0.984} & \textbf{0.953} & \underline{0.946} & \textbf{0.974} & \underline{0.895} \\
ENeRF+Ours & \textbf{0.972} & \underline{0.922} & \underline{0.937} & \underline{0.983} & \textbf{0.953} & 0.942 & \underline{0.972} & \textbf{0.901} \\
\hline
$\text{NeRF}_{10.2h}$ \cite{mildenhall2020nerf} & 0.971 & \underline{0.943} & \textbf{0.969} & 0.980 & \textbf{0.975} & \textbf{0.968} & 0.981 & \textbf{0.908} \\
$\text{IBRNet}_{ft-1.0h}$ \cite{wang2021ibrnet} & 0.955 & 0.913 & 0.940 & 0.978 & 0.940 & 0.937 & 0.974 & 0.877 \\
$\text{MVSNeRF}_{ft-15min}$ \cite{chen2021mvsnerf} & 0.934 & 0.898 & 0.944 & 0.971 & 0.924 & 0.927 & 0.970 & 0.879 \\
$\text{ENeRF}_{ft-1.0h}$ \cite{lin2022efficient} & 0.971 & \textbf{0.960} & 0.939 & 0.985 & 0.949 & 0.947 & \textbf{0.985} & 0.893 \\
$\text{ENeRF+Ours}_{ft-15min}$ & \underline{0.974} & 0.936 & 0.948 & \underline{0.987} & 0.961 & 0.956 & \underline{0.983} & \underline{0.907} \\
$\text{ENeRF+Ours}_{ft-1.0h}$ & \textbf{0.976} & 0.938 & \underline{0.950} & \textbf{0.988} & \underline{0.963} & \underline{0.959} & \textbf{0.985} & \textbf{0.908} \\
\hline
Metric & \multicolumn{8}{c}{LPIPS} \\
\hline
PixelNeRF \cite{yu2021pixelnerf} & 0.386 & 0.421 & 0.335 & 0.433 & 0.427 & 0.432 & 0.329 & 0.526 \\
IBRNet \cite{wang2021ibrnet} & 0.144 & 0.241 & 0.159 & 0.175 & 0.202 & 0.164 & 0.103 & 0.369 \\
MVSNeRF \cite{chen2021mvsnerf} & 0.282 & 0.187 & 0.211 & 0.173 & 0.204 & 0.216 & 0.177 & 0.244 \\
ENeRF \cite{lin2022efficient} & 0.055 & 0.110 & 0.076 & 0.059 & 0.075 & 0.084 & 0.039 & 0.183 \\
MatchNeRF \cite{chen2023matchnerf} & 0.107 & 0.185 & 0.117 & 0.162 & 0.160 & 0.119 & 0.060 & 0.398 \\
GNT \cite{wang2023attention} & 0.065 & 0.116 & \textbf{0.063} & 0.095 & 0.112 & \textbf{0.025} & 0.243 & \textbf{0.115} \\
MVSGaussian \cite{liu2024mvsgaussian} & \underline{0.036} & \textbf{0.091} & \underline{0.069} & \textbf{0.040} & \textbf{0.066} & \underline{0.063} & \textbf{0.027} & \underline{0.179} \\
ENeRF+Ours & \textbf{0.034} & \underline{0.104} & 0.073 & \underline{0.051} & \underline{0.072} & 0.075 & \underline{0.029} & 0.184 \\
\hline
$\text{NeRF}_{10.2h}$ \cite{mildenhall2020nerf} & 0.055 & 0.101 & \textbf{0.047} & 0.089 & 0.054 & 0.105 & 0.033 & 0.263 \\
$\text{IBRNet}_{ft-1.0h}$ \cite{wang2021ibrnet} & 0.079 & 0.133 & 0.082 & 0.093 & 0.105 & 0.093 & 0.040 & 0.257 \\
$\text{MVSNeRF}_{ft-15min}$ \cite{chen2021mvsnerf} & 0.129 & 0.197 & 0.171 & 0.094 & 0.176 & 0.167 & 0.117 & 0.294 \\
$\text{ENeRF}_{ft-1.0h}$ \cite{lin2022efficient} & 0.030 & \textbf{0.045} & 0.071 & 0.028 & 0.070 & 0.059 & \textbf{0.017} & 0.183 \\
$\text{ENeRF+Ours}_{ft-15min}$ & \underline{0.028} & 0.082 & 0.061 & \underline{0.027} & \underline{0.053} & \underline{0.048} & 0.019 & \textbf{0.175} \\
$\text{ENeRF+Ours}_{ft-1.0h}$ & \textbf{0.026} & \underline{0.077} & \underline{0.059} & \textbf{0.025} & \textbf{0.048} & \textbf{0.044} & \underline{0.018} & \underline{0.179} \\
\hline
\end{tabular}
\caption{Quantitative results of other eleven scenes on the NeRF Synthetic \cite{mildenhall2020nerf} test set. The best result is highlighted in bold, while the second-best is underlined.}
\label{tab:per-scene-nerf}
\end{table*}

As shown in \cref{tab:per-scene-dtu}, \cref{tab:per-scene-other-dtu} and \cref{tab:per-scene-nerf}, ENeRF+Ours achieves the best generalizable performance on most scenes. For per-scene optimization results, ENeRF+Ours ranks second only to NeRF \cite{mildenhall2020nerf} in terms of PSNR, while achieving the best performance in terms of SSIM and LPIPS. A likely explanation is that our loss function incorporates SSIM loss and perceptual loss in addition to MSE. This combination enables better preservation of high-frequency details in the images while enhancing perceptual quality.

\cref{tab:per-scene-llff} demonstrates that ENeRF+Ours consistently ranks either first or second in both generalizable and per-scene optimization results across all scenes, ranking first on most. This further validates the effectiveness of our method, particularly in natural scenes characterized by piece-wise smoothness.

\section{Limitations}
Our method leverages depth information to guide sampling, primarily relying on MVS depth estimation. However, its effectiveness can be compromised in scenes where MVS performance is poor. For instance, in the Ficus and Materials scenes from the NeRF Synthetic dataset, challenges such as occlusion, thin structures, reflections, or glossy surfaces result in poor MVS depth estimation. Consequently, our generalizable results are not as strong as GNT \cite{wang2023attention}, which is based on an attention mechanism, and our per-scene optimization results fall short of NeRF \cite{mildenhall2020nerf} due to fewer samples that cannot be accurately placed near surfaces.

\begin{figure}[!ht]
  \centering
  \subfloat[Target View]{
    \includegraphics[width=0.47\columnwidth]{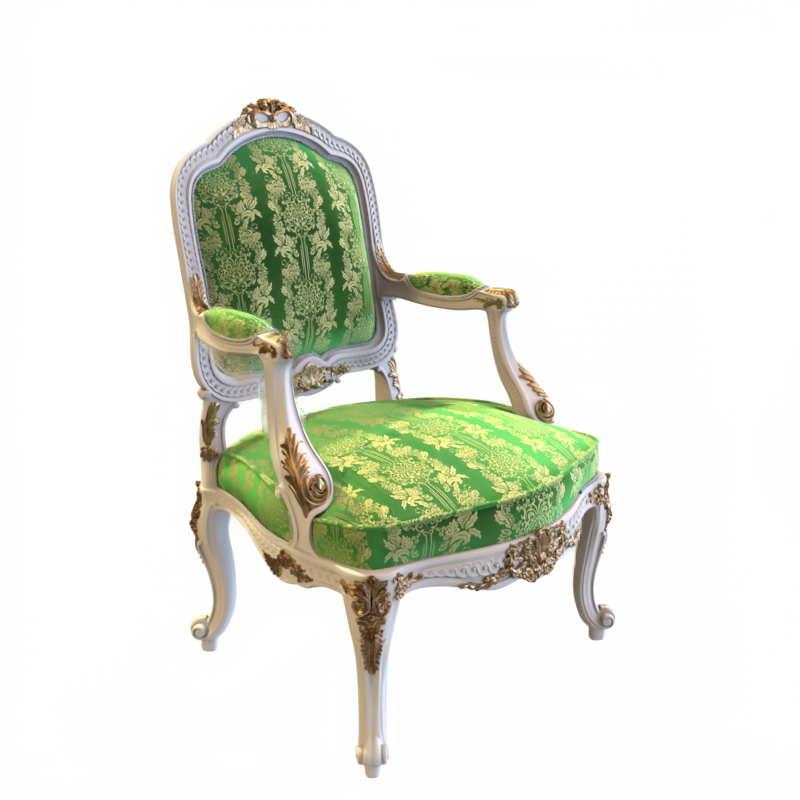}
  }
  \subfloat[Sample Distribution]{
    \includegraphics[width=0.47\columnwidth]{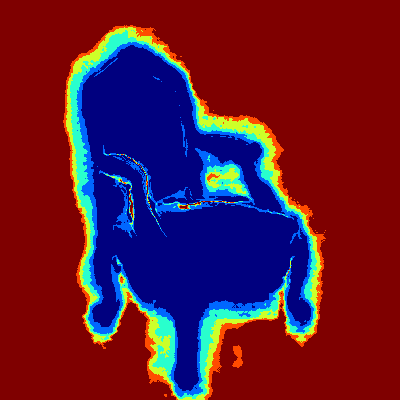}
  }
  \caption{Visualization of sample allocation in depth-guided adaptive sampling.}
  \label{fig:object-centric-smpl}
\end{figure}

Additionally, the efficiency of our method faces challenges on object-centric datasets, such as the NeRF Synthetic dataset or certain human datasets. In these cases, the lack of background hinders depth estimation, leading our method to allocate the maximum number of samples in these regions. This misallocation reduces the overall efficiency of our approach, as illustrated in \cref{fig:object-centric-smpl}. However, this issue can be mitigated by employing a mask to exclude these areas from sampling.

For MVSGaussian+Ours, our method is applied to the NeRF branch of MVSGaussian. Since the NeRF branch in MVSGaussian uses only a single sample per ray, the speed improvement achieved by our approach is relatively small. One reason our bundle sampling enhances rendering quality is its ability to exploit correlations between adjacent rays. However, in MVSGaussian, a post-processing network is employed to leverage the context between adjacent 3D-GS elements, which partially diminishes the effectiveness of our method. Exploring more effective ways to integrate our approach with 3D-GS will be the focus of our future research.

{
    \small
    \bibliographystyle{ieeenat_fullname}
    \bibliography{main.bib}
}

\captionsetup{font=small}
\captionsetup[table]{aboveskip=0pt}
\captionsetup[table]{belowskip=0pt}
\newcommand{\toreviewer}[1]{\vspace{0.4em}\noindent \textcolor{blue}{\textbf{#1 \hspace{0.2em}}}}
\newcommand{\issue}[1]{\vspace{0.1em}\noindent \textbf{#1 \hspace{0.2em}}}

\clearpage
\setcounter{page}{1}
\setcounter{section}{1}

We sincerely thank all reviewers for their constructive feedback on our paper. We are encouraged that our work has been recognized as well-motivated (Reviewer JZVb, iQq1), making meaningful contributions (Reviewer JZVb, 7QT7), with well-executed ablation studies (Reviewer JZVb, iQq1), tackles an important problem (7QT7), and is both effective (Reviewer iQq1, 7QT7) and efficient (Reviewer 7QT7). Below, we provide a point-by-point response to the reviewers’ comments. We hope this addresses your concerns and helps you consider recommending our paper for acceptance.


\toreviewer{To Reviewer JZVb:}

\issue{Input view setup}
We use 2 and 3 input views following ENeRF and MVSGaussian to ensure fair comparisons. \cref{tab:10-view} lists the results with 10 input views, where all 3 methods are re-trained. For efficiency, the bundle size is fixed at $2 \times 2$. The results indicate that additional input views provide limited improvement, as these methods rely on depth information, a common strategy for few-shot NeRF.

\begin{table}[h]
\centering
\vspace{-1em}
\resizebox{0.48\textwidth}{!}{
\Large
\begin{tabular}{cccccccccc}
\hline
\multirow{2}{*}{Methods} & \multicolumn{3}{c}{DTU} & \multicolumn{3}{c}{Real Forward-facing} & \multicolumn{3}{c}{NeRF Synthetic} \\
\cmidrule(r){2-4} \cmidrule(r){5-7} \cmidrule(r){8-10}
 & PSNR $\uparrow$ & SSIM $\uparrow$ & LPIPS $\downarrow$ & PSNR $\uparrow$ & SSIM $\uparrow$ & LPIPS $\downarrow$ & PSNR $\uparrow$ & SSIM $\uparrow$ & LPIPS $\downarrow$ \\
\hline
ENeRF &  23.95 & 0.947 & 0.117 & 22.60 & 0.836 & 0.223 & 22.82 & 0.905 & 0.245 \\
MVSGaussian & \underline{28.48} & \underline{0.966} & \underline{0.072} & \underline{23.79} & \pmb{0.865} & \underline{0.190} & \underline{26.27} & \pmb{0.948} & \underline{0.094} \\
ENeRF+Ours & \pmb{29.09} & \pmb{0.967} & \pmb{0.067} & \pmb{24.05} & \underline{0.864} & \pmb{0.179} & \pmb{26.37} & \underline{0.947} & \pmb{0.079} \\
\hline
\end{tabular}
}
\caption{Quantitative comparison with 10 input views.}
\label{tab:10-view}
\vspace{-1em}
\end{table}

\issue{Application to methods}
Our method is compatible with various existing approaches but relies on multi-scale features and depth estimation for optimal performance. While most methods incorporate at least one of these components, MVSNeRF and IBRNet lack both, necessitating additional modifications for direct integration with our approach.


\toreviewer{To Reviewer iQq1:}

\issue{Colors from $K \times K$ rays}
Our method imposes no additional constraints on camera orientation. The $K \times K$ rays are emitted from the target view, and samples are reprojected into source views to extract colors. This ensures alignment with the corresponding pixels in the target view.

\issue{Efficiency evaluation} We only evaluate FPS on DTU following ENeRF and MVSGaussian. \cref{tab:FPS} presents FPS results for the Real Forward-Facing and NeRF Synthetic datasets, showing that FPS variations align with resolution changes.

\begin{table}[h]
\centering
\vspace{-1em}
\resizebox{0.48\textwidth}{!}{
\Large
\begin{tabular}{*{10}{c}}
\toprule
    &  & ENeRF & MVSGaussian & ENeRF+Ours ($2 \times 2$) & ENeRF+Ours ($4 \times 4$) \\
    \midrule
    & Real Forward-facing & 14.04 & 15.88 & 19.09 & \pmb{31.02} \\
    & NeRF Synthetic & 12.87 & 13.56 & 17.08 & \pmb{26.68} \\
    \bottomrule
\end{tabular}
}
\caption{Comparison of rendering speed (FPS) with 3 input views.}
\label{tab:FPS}
\vspace{-1em}
\end{table}

\issue{W4 and W6} We have revised follow your suggestion.


\toreviewer{To Reviewer iQq1 and 7QT7:}

\issue{Relation to Plenoptic sampling theory} The Plenoptic sampling theory highlights that the accuracy of the geometry influences the sampling rate needed for image capture. By decomposing a scene into $D$ depth ranges and separately sampling the light field within each range, the camera sampling interval can be increased by a factor of $D$. This inspired the use of multi-plane images (MPIs), which represent a scene as fronto-parallel RGB$\alpha$ planes evenly sampled in disparity within a reference camera’s view frustum. As Ravi Ramamoorthi, the author of local light field fusion and NeRF, noted, MPIs can be seen as discrete volumetric radiance fields sampled at various depth layers, whereas NeRFs represent a continuous volume \cite{ramamoorthi2024sampling}. In this context, samples along each ray in NeRF can be regarded as a local MPI, where each sample corresponds to a local light field within a small disparity range $h_d \le \frac{\text{max} \left( 2 \Delta v, 1 / B_v^s \right)}{f \cdot \Delta t}$.

This understanding inspires our integration of plenoptic sampling theory with NeRF. The maximum camera spacing $\Delta t \le \frac{\text{max} \left( 2 \Delta v, 1 / B_v^s \right)}{f \cdot h_d}$, is determined based on maximum frequency $B_v^s$, camera resolution $\Delta v$, and depth range $h_d$. However, the piecewise smooth nature of natural scenes means high-frequency content is localized, with most regions being smooth. This allows for reduced sampling rates in smooth areas by lowering the camera resolution (increasing $\Delta v$), while maintaining fixed $\Delta t$. We assume $2 \Delta v \le 1 / B_v^s$ to ensure the frequency is captured. However, in high-frequency regions (e.g., edges, textures, thin structures), low-resolution sampling can cause information loss. To address this, we aggregate features from related regions (joint bundle representation) using sphere sampling and supplement high-frequency details with pixel-aligned colors (ray-specific representation). Furthermore, high-frequency regions often suffer from inaccurate depth estimation, necessitating more samples to ensure each sample covers a small disparity range. This, in turn, increases sampling density along epipolar lines in source views, enhancing sampling density in the source views to some extent.

In summary, we extend plenoptic sampling by transitioning from sampling all rays at uniform depths to sampling different bundles at varying depths. Furthermore, our depth-guided adaptive sampling places only as many samples as necessary at the required density, unlike existing methods that typically fix the number of samples. For instance, ENeRF may oversample in smooth areas and undersample in high-frequency or depth-discontinuous regions.

We recognize that our initial explanation was too brief, limiting accessibility for readers. We have improved our statement to make the relationship between plenoptic sampling theory and our method easier to understand.

\issue{Missing references} We have included them.

\toreviewer{To Reviewer 7QT7:}

\issue{Title of Sec. 3} We have revised follow your suggestion.

\issue{Missing explanation} $f_{voxel}$ is adopted from ENeRF. We have added its explanation in the main text.

{
    \small
    \bibliographystyle{ieeenat_fullname}
    \bibliography{main.bib}
}
\end{document}